\documentclass[preprint, 11pt, 3p]{elsarticle}

\usepackage[T1]{fontenc}
\usepackage{amssymb}
\usepackage{bm}
\usepackage{amsmath}
\usepackage[normal]{subfigure}
\usepackage{graphicx}
\usepackage{multirow} 
\usepackage{xcolor,xspace}
\usepackage{colortbl}
\usepackage{float}
\usepackage{hyperref}
\usepackage{algorithm}
\usepackage[noend]{algpseudocode}
\usepackage{natbib}
\usepackage{tabularx} 
\usepackage{booktabs}
\usepackage{amsfonts}
\usepackage{scalefnt}
\usepackage{array}
\usepackage{csvsimple}
\usepackage{pgfplots}
\pgfplotsset{compat=newest}
\usepackage{tikz}
\usepackage{mathabx}
\usetikzlibrary{external}
\tikzset{external/system call={pdflatex \tikzexternalcheckshellescape 
                                        -halt-on-error
                                        -interaction=batchmode 
										--extra-mem-bot=999999999
										--extra-mem-top=999999999
                                        -jobname "\image" "\texsource"
                                        }} 
\usepgfplotslibrary{fillbetween}
\usetikzlibrary{arrows.meta,decorations.pathmorphing,mindmap,backgrounds,arrows,positioning,shapes,calc,patterns}
\DeclareGraphicsExtensions{.pdf}
\DeclareMathAlphabet{\mathsfit}{\encodingdefault}{\sfdefault}{m}{sl}
\SetMathAlphabet{\mathsfit}{bold}{\encodingdefault}{\sfdefault}{bx}{sl}

\newcommand{\stov}[1]{\bm{#1}}
\newcommand{\detv}[1]{\mathbf{#1}}

\newcommand{\detm}[1]{\mathbf{#1}}
\newcommand{\ac}[1]{#1}
\newcommand{\ie}{i.\,e.\;}

\newcommand{\eg}{e.\,g.\;}

\newcommand{\citepeg}[1]{(\eg \cite{#1})}
\DeclareMathOperator*{\argmin}{arg\,min}
\newif\ifusepdf
\usepdftrue

\newcounter{pdffigno}

\makeatletter
\providecommand{\doi}[1]{%
	\begingroup
	\let\bibinfo\@secondoftwo
	\urlstyle{rm}%
	\href{http://dx.doi.org/#1}{%
		doi:\discretionary{}{}{}%
		\nolinkurl{#1}%
	}%
	\endgroup
}
\makeatother

\journal{Applied Soft Computing (DOI: \href{http://dx.doi.org/10.1016/j.asoc.2021.107807}{10.1016/j.asoc.2021.107807})}
\begin{document}
\newcommand{\changemine}[1]{#1}
\newcommand{\changerev}[1]{#1}
\newcommand{\changeown}[1]{#1}
\newcommand{\changefinal}[1]{#1}
\begin{frontmatter}

\title{Local Latin Hypercube Refinement for Multi-objective Design Uncertainty Optimization} 

\author[label1,label2]{Can Bogoclu\corref{cor1}}
\ead{Can.Bogoclu@rub.de}

\author[label2]{Dirk Roos}
\ead{Dirk.Roos@hs-niederrhein.de}

\author[label1]{Tamara Nestorovi\'{c}}
\ead{Tamara.Nestorovic@rub.de}

\address[label1]{Ruhr-Universit\"{a}t Bochum, Faculty of Civil and Environmental Engineering, Institute of Computational Engineering, Mechanics of Adaptive Systems, Universit\"{a}tsstr. 150, Building ICFW 03-725, D-44801 Bochum, Germany}
\address[label2]{Niederrhein University of Applied Sciences, Faculty of Mechanical and Process Engineering, Institute of Modelling and High-Performance Computing, Reinarzstr. 49, D-47805 Krefeld, Germany}

\cortext[cor1]{I am the corresponding author.}

\begin{abstract}
\changerev{Optimizing the reliability and the robustness of a design is important but often unaffordable due to high sample requirements. Surrogate models based on statistical and machine learning methods are used to increase the sample efficiency. However, for higher dimensional or multi-modal systems, surrogate models may also require a large amount of samples to achieve good results. We propose a sequential sampling strategy for the surrogate based solution of multi-objective reliability based robust design optimization problems. Proposed local Latin hypercube refinement (LoLHR) strategy is model-agnostic and can be combined with any surrogate model because there is no free lunch but possibly a budget one. The proposed method is compared to stationary sampling as well as other proposed strategies from the literature. Gaussian process and support vector regression are both used as surrogate models. Empirical evidence is presented, showing that LoLHR achieves on average better results compared to other surrogate based strategies on the tested examples.}
\end{abstract}


\end{frontmatter}


\newlength\fheight 
\newlength\fwidth
\section{Introduction}\label{sec:intro}
\changerev{Design optimization is essential in modern engineering, especially when designing products for well established markets. Most designs have several quality indicators such as the production costs, operational costs, performance, reliability and robustness. Thus, design optimization problems in industrial applications are often multi-objective (MO) and some of these objectives are often competing. Moreover, the last two indicators relate to the aleatoric uncertainties, that may arise through production tolerances, operational variations and other irreducible sources.}

\changerev{Various MO optimization algorithms were proposed using evolutionary \citepeg{PAES1999,NSGAII2002} or deterministic strategies \citepeg{Das1998,Lovison2021}. All of these algorithms require several evaluations of the indicator functions and the latter also require their gradients. Such indicators are often computed using expensive numerical simulations or physical experiments. Moreover, these indicator functions and their gradient may be unavailable or intractable due to the complexity of the physical system. In this case, the gradient and the uncertainty assessment have to be conducted using sampling based strategies, requiring multiple function evaluations (\ie samples) for each candidate design during the optimization. Hence, surrogate models are often deployed for expensive functions to make the solution affordable \citepeg{Forrester2008a}.}

\changerev{However, surrogate models may also require a high number of samples depending on the problem. Recent research \citepeg{Parr2012,Yang2019} has shown that sequential sampling can increase the sample efficiency of the solution for MO problems without aleatoric parameter uncertainty. Nonetheless, if it is not possible to design a cheaper product with higher performance, robustness and reliability are the only selling points to beat the competition.}

\changerev{Robustness of a design is related to the variation of the performance indicators (often the objectives) under parameter uncertainty \citep[][]{Taguchi1986,Arvidsson2008} and a higher robustness corresponds to a smaller variation. Reliability describes the probability of the design fulfilling some criteria (often the feasibility of the constraints) under parameter uncertainty \citep[][]{Hasofer1974,Kiureghian1990}.}

Various strategies were proposed for the solution of reliability-based (RBDO) \citepeg{Zou2006,Aoues2010,Lopez2012,Yi2016,Stromberg2017,Li2020,Yang2021} and robust design optimization (RDO) \citepeg{Chen1996,Hwang2001,Koch2004} problems, which seek to reduce the number of required samples for the uncertainty assessment. Specifically, instead of \textit{nesting} the uncertainty quantification in the optimization loop as proposed by \eg \cite{Bichon2009}, they \textit{decouple} the uncertainty assessment and solve a series of deterministic problems to achieve the desired level of reliability and robustness. However, the number of required samples is still relatively high for expensive black-box functions.

For increasing the sample efficiency, surrogate based strategies using stationary (\ie one-shot) sampling for RBDO \citepeg{Bucher1990,Yu2002,Du2004,Allen2004,Youn2004_2,Su2014,Mansour2016} as well as for RDO \citepeg{Enevoldsen1994,Akkerman2000,Chakraborty2017} were proposed before. \changerev{Moreover, sequential sampling strategies were shown to increase the sample efficiency further in RBDO \citepeg{Dubourg2011,Zhuang2012,Lopez2017,Moustapha2016,Zhang2019,Wauters2021} and RDO \citepeg{Sun2014} problems.}

All of these sequential sampling strategies are formulated for a specific surrogate choice; usually Kriging \citep[][]{Krige1951} also called the Gaussian process (GP) \citep[][]{Rasmussen2006}. \changerev{However, as the \textit{no free lunch} theorem \citep{Wolpert1996} states in simple words, there is no single model which outperforms other models in all data sets. Despite the clever use of probabilistic model information in some works, the approximation quality of the model is more important regarding the results and the best quality may be achieved by a non-probabilistic model in some cases. Thus, decoupling the choice of SML model from the proposed method without sacrificing the advantages of sequential refinement schemes is one of the motivations behind this work.}


\changerev{Finally, there is also previous research \citepeg{Youn2005,Lee2008,Antonio2009,Martowicz2012,Yu2013,Motta2016,Zhang2018,Leimeister2021} regarding the reliability-based robust design optimization (RRDO), a combination of RBDO and RDO but all works listed above investigate single-objective (SO) problems. Although the MORBDO \citepeg{Sinha2007,Zhao2007,Fang2013,Rashki2014,Tsoukalas2015,Sun2017,Duan2019,Lim2020,Feng2021} and MORDO \citepeg{Rangavajhala2013,Fang2015,Li2015,Yu2017,Groetzner2021} are also actively researched, the number of publications regarding the MORRDO is relatively small.}

\changerev{Strategies proposed in \cite{Yadav2010,Shahraki2014,Liu2014,Lobato2020} do not use surrogate models and thus report a high number of function evaluations. \cite{Gu2013} apply deterministic optimization, \ac{MORDO}, \ac{MORBDO} and \ac{MORRDO} to a vehicle crashworthiness design using a surrogate based sequential sampling strategy and radial basis functions. \cite{Gu2014} apply the same \ac{MORRDO} strategy using support vector regression (SVR) to the design optimization of an occupant restraint system and compare the results with the deterministic solution. \cite{Yang2014} propose a surrogate based strategy using the moment based \ac{RRDO} formulation. A pure quadratic polynomial is used as the surrogate model. Accuracy of this model for non-polynomial functions or functions of higher order may decrease, yielding suboptimal solutions. Since the type and the degree of relationships are assumed to be unknown regarding black-box problems, polynomials may not always be the best choice.  
\cite{Zhuang2015} propose a sequential sampling strategy using \ac{GP}, where multiple objectives are linearly combined to form a sequence of SO problems using varying weights to obtain parts of the Pareto frontier. 
}

\changerev{In any case, beside the sparsity of the publications, all surrogate based strategies for \ac{MORRDO} use a single surrogate method, while the strategies not using surrogate models report a high number of samples. Thus, this work proposes a model agnostic surrogate based sequential sampling strategy that aims to overcome the limitations of a single surrogate model and solve \ac{MORRDO} problems efficiently. Specifically, we propose a strategy that aims to increase the sample efficiency compared to direct optimization, the accuracy and performance of the results compared to stationary sampling. Furthermore, it aims to achieve comparable or better results to methods using sequential sampling with a single surrogate model. Moreover, the proposed strategy should be usable with any optimization algorithm that is capable of solving the underlying problem, since \textit{no free lunch} theorem is also formulated for optimization \citep{WoMac1997}.}

\changerev{The paper is structured as follows: Section \ref{sec:Theo} starts with a formal definition of the \ac{MORRDO} problem, followed by brief descriptions of used methods for surrogate modelling, data generation and clustering. Section \ref{sec:proposal} describes the proposed local Latin hypercube refinement (LoLHR) strategy in detail. Section \ref{sec:results} compares LoLHR to some other baseline algorithms as well as some previously proposed methods  and the results are discussed. In Section \ref{sec:application}, the proposed method is tested on a lead screw optimization problem. Finally in Section \ref{sec:conc}, concluding remarks about the results of this work as well as their implications to future work are included. The implementation details of all algorithms are given in \ref{app:impldetails} for improving reproducibility.}

\section{Theoretical Background}\label{sec:Theo}

\subsection{Multi-Objective Design Optimization}\label{sec:DO}

Constrained MO parametric design optimization problems can be denoted as
\begin{equation}
\begin{aligned}
\argmin_{\detv{x}} \; &f_k(\detv{x}) \qquad &k \in  [1,...,n_o]\\
\textit{\footnotesize s.t.} \quad &c_j(\detv{x}) \geq 0  \qquad &j \in [1,...,n_c]\\
&\detv{x}^l \leq \detv{x} \leq \detv{x}^u&\\
\end{aligned}\label{eq:MOOConst}
\end{equation}
stating that all $n_{o}$ objective functions $f_k(\cdot): \mathbb{R}^n \rightarrow \mathbb{R}$ are to be minimized by finding the optimal values for $n$ input or design parameters $\detv{x} \in \mathbb{R}^n$, which do not violate any of the $n_{\mathrm{c}}$ constraint functions $c_j(\cdot)$. $\detv{x}^l \in \mathbb{R}^n$ and  $\detv{x}^u \in \mathbb{R}^n$ represent the lower and the upper bounds for the search space. 

The solution of problems with multiple competing objectives $n_o > 1$ is a set of points, often referred to as the Pareto frontier. \changerev{For any Pareto optimal design, each objective is optimized to an extent, that an improvement of an objective cannot be achieved without worsening at least one of the other objectives, \ie the design is \textit{non-dominated}.} 
Assuming minimization, a design $\detv{x}^*$ dominates another ($\detv{x}^* \succ \detv{x}$) if the following equations hold.
\begin{equation}
\begin{aligned}
&\forall \quad i \in  [1,\dots,n_o] \quad f_i(\detv{x}^*) \leq f_i(\detv{x})  \\
&\exists \quad k \in [1,\dots,n_o] \quad f_k(\detv{x}^*) < f_k(\detv{x}) \\
&\forall \quad j \in  [1,\dots,n_c] \quad c_j(\detv{x}^*) \geq 0 \qquad  c_j(\detv{x}) \geq 0
\end{aligned}\label{eq:ParetoOptimality}
\end{equation}

In this work, \ac{NSGA}-II as proposed by \citep{NSGAII2002} is used for global MO optimization, which uses the genetic algorithm and is therefore a stochastic and gradient-free programming method. 
\changerev{The choice of the optimization algorithm depends on the problem and the algorithm is interchangeable within the proposed method as long as it can handle MO problems. 
Nonetheless, \ac{NSGA}-II seemed to work well on the investigated problems. Moreover, using such an off-the-shelf algorithm is useful for evaluating the efficiency of proposed strategy. It demonstrates that greedier algorithms can become affordable when used with surrogate models.} 

\subsection{Design Uncertainty Optimization}\label{sec:RRDO}
\subsubsection{Robust Design Optimization}
Robustness of a design is often measured by the variance of the objectives under aleatoric input uncertainty \citepeg{Taguchi1986, Sandgren2002}. \changerev{Quantile-based \citepeg{Razaaly2020} formulations of robustness also exist but since both measure the deviation from the nominal or the mean value, the more traditional formulation is preferred here.} The \ac{MORDO} problem can be denoted as
\begin{equation}
\begin{aligned}
\argmin_{\boldsymbol{\theta}_{\stov{X}}} \quad  &\mathrm{E}^{\boldsymbol{\theta}_{\stov{X}}}\left[f_k(\stov{X})\right],\operatorname{Var}^{\boldsymbol{\theta}_{\stov{X}}}\left[f_k(\stov{X})\right]  \qquad &k \in  [1,...,n_o]\\
\textit{\footnotesize s.t.} \quad &c_j(\boldsymbol{\theta}_{\stov{X}}) \geq 0  \qquad &j \in [1,...,n_c]\\
&\stov{X} \sim F_{\stov{X}}(\cdot \; ; \boldsymbol{\theta}_{\stov{X}})&\\
&\boldsymbol{\theta}_{\stov{X}}^l \leq \boldsymbol{\theta}_{\stov{X}} \leq \boldsymbol{\theta}_{\stov{X}}^u& \\
\end{aligned} \label{eq:RDO}
\end{equation}
where the operators $\operatorname{E}^{\boldsymbol{\theta}_{\stov{X}}}[\cdot]$ and $\mathrm{Var}^{\boldsymbol{\theta}_{\stov{X}}}[\cdot]$ describe the expectation value and the variance operations with respect to the distribution parameters $\boldsymbol{\theta}_{\stov{X}} \in \mathbb{R}^{n_\theta}$ including but not limited to $\boldsymbol{\mu}_{\stov{X}}$, yielding $2 n_o$ objectives from $n_o$ deterministic ones. 

The \changemine{design} parameters are subject to \changemine{aleatoric} uncertainty as represented by the continuous multidimensional random variable $\stov{X} \sim F_{\stov{X}}$ following the cumulative density function (CDF) $F_{\stov{X}}(\cdot \; ; \boldsymbol{\theta}_{\stov{X}}): \mathbb{R}^n \rightarrow \mathbb{R}$. Hence, the objective value $Y_k = f_k(\stov{X})$ is also uncertain although the relationship (\ie the objective function $f_k$) is deterministic. The variance $\operatorname{Var}^{\boldsymbol{\theta}_{\stov{X}}}\left[Y_k\right]$ wrt. the distribution $F_{\stov{X}}$ of the model parameters $\stov{X}$ is to be minimized along with the expectation value to increase the robustness of the design. It is also possible to combine the mean and the variance into a \ac{SO} since the designer may not be interested in solutions that only optimize one of the two or for different formulations such as the quantile-based one.

\changerev{Expectation and variance of the objectives could be computed analytically given the function $f_k(\cdot)$ but this may become intractable for complicated functions and distributions. Sampling based procedures are used for an approximation instead \citep{McKay1979}, which allow $f_k$ to be treated as a black-box function. In this work an orthogonal sampling \citep{Tang1993} is used for the approximation of these moments, which was optimized using the algorithm in \citep{Joseph2008}.}

Note that uncertain operational (\ie non-design, uncontrollable) variables such as material constants may also exist. For a more concise notation, these are included in $\stov{X}$ with fixed distributions in contrast to the design parameters.

\subsubsection{Reliability-based Design Optimization}
For assessing and improving the design reliability, constraints $c_j$ are replaced with the corresponding limit state functions $g_j$ and a probability of failure $P(\mathcal{F}) = P \left(g(\stov{X}) < 0 \right)$ is calculated. For continuous $\stov{X}$, this requires a multidimensional integration over the failure domain. This integral can be denoted for a single limit state function $g$ as
\begin{equation}%
P(\mathcal{F}) = \mathop{\int \mathop{\ldots}^{n} \int}_{g(\detv{x}) < 0} f_{\stov{X}}(x_1, \dots, x_n) d x_1 \dots d x_n 
\label{eq:Pf}%
\end{equation}%
where $f_{\stov{X}}$ denotes the joint probability density function. Reducing or limiting the failure probability $P(\mathcal{F})$ of a design during the optimization is referred to as RBDO \citep{Aoues2010,Dubourg2011,Stromberg2017}. Reliability in this context is often measured as the probabilistic inverse of the failure probability $1 - P(\mathcal{F})$ or the corresponding $\beta$-level safety index $-\Phi^{-1}(P(\mathcal{F}))$, where $\Phi^{-1}$ denotes the CDF of a standard normal variable.

Solving the integral in Eq. \ref{eq:Pf} is not trivial and often a burdensome process. Various sampling methods \citepeg{Fu1994-1,Bucher2015,Lee2017} are proposed for this task, each having different strengths and weaknesses depending on the problem. In this work, Directional sampling (DS) \citep{Bjerager1988,Ditlevsen1988} and Monte-Carlo simulation (MC) \citep{Rubinstein1981a} were preferred because of their accuracy with multimodal limit state functions. 

A \ac{MORBDO} problem can be denoted as
\begin{equation}
\begin{aligned}
\argmin_{\boldsymbol{\theta}_{\stov{X}}} \; &f_k(\boldsymbol{\theta}_{\stov{X}}) \qquad  k \in  [1,...,n_o]  \\
\textit{\footnotesize s.t.} \quad &P(\mathcal{F}) \leq P^t(\mathcal{F}) \\
&\stov{X} \sim F_{\stov{X}}(\cdot \; ; \boldsymbol{\theta}_{\stov{X}}) \\
&\boldsymbol{\theta}_{\stov{X}}^l \leq \boldsymbol{\theta}_{\stov{X}} \leq \boldsymbol{\theta}_{\stov{X}}^u \\
\end{aligned} \label{eq:RBDO}
\end{equation}

The above formulation considers the combined probability of failure $P(\mathcal{F})$ of a \textit{series} system, where all $g_j(\stov{X})$ are evaluated simultaneously.
\begin{equation}
P(\mathcal{F}) = P\left(\min_j\left(g_j(\stov{X})\right) < 0\right)  \qquad j \in \left[1,...,n_c\right] \label{eq:FailProb}
\end{equation}
For some applications, it may be desirable to set different target probabilities to each limit state function $g_j$. Although such cases are not investigated in this work, the applicability of the proposed method does not depend on the formulation of the investigated probabilities. 

Finally, $P(\mathcal{F})$ can also be used as an objective to be optimized instead of or in addition to the constraint $P(\mathcal{F}) \leq P^t(\mathcal{F})$. Although these cases are not explicitly denoted here, an example is given later in empirical results along with the corresponding notation.

\subsubsection{Reliability-based Robust Design Optimization}
The combination of \ac{RDO} and \ac{RBDO} with limiting failure probability $P^t(\mathcal{F})$ is called \ac{RRDO} \citep{Youn2008,Tang2012} and can be denoted as
\begin{equation}
\begin{aligned}
\argmin_{\boldsymbol{\theta}_{\stov{X}}} \;  &\mathrm{E}^{\boldsymbol{\theta}_{\stov{X}}}\left[f_k(\stov{X})\right],\operatorname{Var}^{\boldsymbol{\theta}_{\stov{X}}}\left[f_k(\stov{X})\right] \qquad &k \in  [1,...,n_o]\\
\textit{\footnotesize s.t.} \quad &P(\mathcal{F}) \leq P^t(\mathcal{F})&\\
&\stov{X} \sim F_{\stov{X}}(\cdot \; ; \boldsymbol{\theta}_{\stov{X}}) \\
&\boldsymbol{\theta}_{\stov{X}}^l \leq \boldsymbol{\theta}_{\stov{X}} \leq \boldsymbol{\theta}_{\stov{X}}^u \\
\end{aligned} \label{eq:StochOpt}
\end{equation}
The proposed strategy in this work seeks to solve the problems in this form, as both robustness and reliability are important for a high quality design.

\subsection{Surrogate Modelling}\label{sec:ML}
For replacing the expensive numerical or physical experiments, surrogate models are often used in engineering applications to increase the sample efficiency of the further analysis such as optimization \citep{Forrester2008a}. For real valued scalar inputs $\detv{x} \in \mathbb{R}^n$ and output $y \in \mathbb{R}$, a surrogate model approximates the relationship $\tilde{f}(\detv{x}) = \tilde{y}$ based on the input data $\detv{X}^0 \in \mathbb{R}^{m \times n}$ as well as the corresponding system or model response $\detv{y}^0 \in \mathbb{R}^m$. $m$ and $n$ denote the number of samples and the number of input parameters, respectively. \changerev{Methods for surrogate modelling are called regression, response surface, supervised machine learning \ac{SML}, statistical learning or probabilistic learning depending on the method, the domain and the context they are used in.} 

In this work, kernel based methods are used due to their efficiency with low data and multimodal functions. A kernel $k(\detv{x}_1, \detv{x}_2): \mathbb{R}^{2n} \rightarrow \mathbb{R}$ can be interpreted as a measure of similarity between two points $\detv{x}_1, \detv{x}_2$. \changerev{Stationary kernels used in this work are defined using some distance metric such as the Euclidean distance $r = ||\detv{x}_1 - \detv{x}_2||_2$, where $||\cdot||_2$ denotes the $l_2$ norm.}

\subsubsection{Kriging and Gaussian Process}
A Gaussian process (\ac{GP}) \citep{Rasmussen2006} is a collection of random variables, any finite number of which have a joint Gaussian distribution. Thus, any prediction point $Y_p$ is also a Gaussian variable with a defined mean $\mu_{Y_p} = \mu_Y(\detv{x})$ and variance $\sigma^2_{Y_p} = \sigma^2_Y(\detv{x})$. Idea of using a \ac{GP} for function approximation emerged from the field of geostatistics under the name of Kriging \citep{Krige1951}. 

The prediction function $\tilde{f}(\cdot)$ of a GP with zero process mean is defined as
\begin{equation}
\begin{aligned}
\tilde{f}(\detv{x}_p) &= \mu_{Y}(\detv{x}) = \detv{k}(\detm{K}+\sigma_\epsilon^2 \detm{I}_m)^{-1}\detv{y}^0 \\
\detv{k} &= \begin{bmatrix}k_1, & \dots, & k_m \end{bmatrix} \\
k_i &= k(\detv{x}_p,\detv{x}^0_{i}; \boldsymbol{\theta}^h, \sigma^2)\\
\detm{K}_{:,i} &= \detm{K}_{i,:}^T = \detv{k}(\detv{X}^0_{i,:},\detm{X}^0; \boldsymbol{\theta}^h, \sigma^2)
\label{eq:GP}
\end{aligned}
\end{equation}
where $\sigma_\epsilon$ is the process or measurement noise as represented in the response data $\detv{y}^0$  and $\detv{x}^0_i = \detm{X}^0{i,:}$ is the $i$-th sample of the observed input points $\detm{X}^0$. $\detm{K}_{:, i}$ denotes the $i$-th row of the covariance matrix $\detm{K}$. Beside $\sigma_\epsilon$, the vector of kernel parameters $\boldsymbol{\theta}^h \in \mathbb{R}^n$ along with the constant process variance $\sigma^2$ are obtained through \textit{training}, \ie optimization of some loss function. To this end, type-II maximum likelihood estimation is used in this work  \citep[see][]{Rasmussen2006}, where the log-likelihood $\mathcal{L}$ of the parameters
\begin{equation}
\begin{aligned}
\mathcal{L}(\boldsymbol{\theta}^h, \sigma, \sigma_\epsilon) = &-\frac{m}{2} \mathrm{log}(2\pi) - \frac{\mathrm{log}\left(\left|\detm{K} +\sigma_\epsilon^2  \detm{I}_m\right|\right)}{2} -\frac{\detv{y}^{0, T} \left(\detm{K}+\sigma_\epsilon^2 \detm{I}_m\right)^{-1}\detv{y}^{0}}{2} \label{eq:GPLikelihood}
\end{aligned}
\end{equation}
were maximized. See \ref{app:sml} for further details.

\subsubsection{Support vector regression}
Support vector regression (SVR) \citep{Cortes1995, SmoScho2003} is also a kernel based method which can be used for surrogate modelling \citep{Tan2011}. Some comparison studies such as \citep{LiNgXieGoh2010} report SVR to sometimes achieve better results than GP. Although it lacks a probabilistic interpretation, SVR was the preferred tool for handling larger data sets before the rise of deep learning paradigm, since SVR only requires a subset of the data for the approximation and can thus scale better to larger data sets compared to standard GP. The prediction function of SVR is given as
\begin{equation} 
\tilde{f}(\detv{x}) = \sum\limits_{i}^m c_i k(\detv{x},\detv{x}^0_i;\theta^h)  + b 
\end{equation}
where $k$ denotes the kernel function. Thus, the prediction is a linear combination of the kernel evaluations $k(\detv{x},\detv{x}^0_i;\theta^h)$. Note that, if both use the same kernel, SVR could result in the same model as the mean GP given in Eq. \ref{eq:GP} if $\detv{c} = (\detm{K}+\sigma_\epsilon^2 \detm{I}_m)^{-1}\detv{y}^0$.

However, $\detv{c}$ is obtained quite differently than GP. Instead of computing $c_i$ by maximizing the likelihood as in GP or minimizing the least squares error as in radial basis functions \citep{Simonoff1996}, they are found by solving the following optimization problem in $\epsilon$-SVR used in this work:
\begin{equation}
\begin{aligned}
\argmin_{\detv{c},\boldsymbol{\zeta}_u,\boldsymbol{\zeta}_l} \;  &\frac{1}{2} || \detv{c} ||_2^2 + \lambda \sum\limits_{i}^m (\zeta_{i,u} + \zeta_{i,l}) \\
\textit{\footnotesize s.t.}: \quad &y_i^0 - \tilde{f}(\detv{x}) \leq \epsilon + \zeta_{i,l} \\
&\tilde{f}(\detv{x}) -  y_i^0  \leq \epsilon + \zeta_{i,u}\\
&\zeta_{i,u} \geq 0 \qquad \zeta_{i,l}  \geq 0 \qquad  i \in [1, m]
\end{aligned} \label{eq:svropt}
\end{equation}
$\zeta_{i,u}$ and $\zeta_{i,l}$ are slack variables in Eq. \ref{eq:svropt} that relax the constraints. $\epsilon$ defines a tolerance margin around the prediction curve for a smoother approximation, \ie errors smaller than $\epsilon$ are ignored. $\lambda$ is the penalty term for the violation of the constraints, \ie when the observations lie beyond the $\epsilon$-margin. Details of training procedure to obtain $\epsilon$ and $\lambda$ are given in \ref{app:sml}.

\subsection{Latin Hypercube Sampling}
\changerev{Latin Hypercube Sampling (LHS) \citep{McKay1979} is a stratified Monte Carlo sampling method and a special case of orthogonal sampling \citep{Tang1993}, where the marginal distribution is uniform.  First, the parameter space is partitioned in regions of equal probability called bins. In the case of uniform distribution, all of the bins are $n$-dimensional hypercubes of equal size. Subsequently, the bins are filled with randomly chosen points, so that each bin is occupied only once for each dimension. This ensures that the desired marginal distribution is achieved while improving the space-filling properties. Switching the values of the resulting sample matrix along each dimension also results in a Latin hypercube design with the same marginal distribution.}


\subsection{Clustering}
\changerev{Clustering is an unsupervised machine learning (UML) task for finding groups of \textit{similar} samples within a data set according to some predefined similarity definition. Density-based spatial clustering of applications with noise (\ac{DBSCAN}) \citep{DBSCAN} is used in this work for clustering especially due to its ability for handling non-convex clusters. It can find arbitrarily shaped clusters by analysing the pairwise distances between the points and thus the sampling density. The number of clusters are inferred automatically, given the $d_{\min}$ which defines the distance of two points to be considered neighbours as well as the number of minimum points $n_k$ to build a cluster.}

\changerev{Especially $d_{\min}$ has a large influence on the resulting clusters. Smaller values tend to generate more clusters and vice versa. Using \ac{DBSCAN} can result in points labelled as noise which could not be assigned to any cluster. This is controlled by both $d_{\min}$ and $n_k$. Nonetheless, small values of $n_k$ do not have a large influence and $d_{\min}$ can be found heuristically, \eg by iterating over the percentile values of the pairwise distances as described in \ref{app:uml}.}

\section{Local Latin Hypercube Refinement for Uncertainty Optimization}\label{sec:proposal}
\begin{figure}[H]
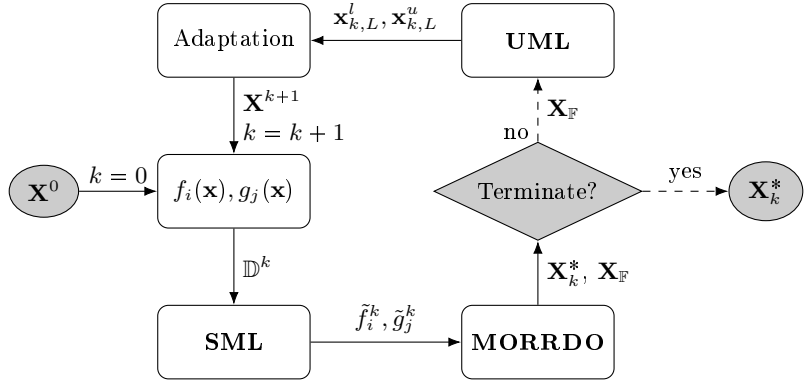

	\centering
		%
\ifusepdf%
	\includegraphics[]{Manuscript-figure\thepdffigno.pdf}%
	\refstepcounter{pdffigno}%
\else%
	\input{./Pictures/FlowChart_Adaptivity.tex}%
\fi%

	\caption{The flow diagram for the proposed framework. Except for the training data $\detm{X}^{k}$ and $\detm{Y}^{k}$, an evaluation of the expensive response functions is not required. Note that each sample in  $\detm{X}^{k}$ is evaluated once and the results are recorded to be used in following iterations.}
	\label{fig:FlowChartAdapt}
\end{figure}

As the name suggests, \ac{LoLHR}\footnote{Code Ocean: \href{https://doi.org/10.24433/CO.2162059.v1}{10.24433/CO.2162059.v1}, github: \href{https://github.com/canbooo/duqo}{canbooo/duqo}} seeks to sequentially refine the surrogate model in local regions, that are thought to be important for \ac{MORRDO} tasks, \ie the neighbourhood of the Pareto frontier. Due to uncertainty, the region of interest encloses but is not limited to the points in the Pareto frontier. Through this refinement, it is expected that surrogate models achieve a better accuracy in these local regions. The idea for the refinement was first described in the context of reliability assessment in \cite{Bogoclu2017}. A flow diagram of \ac{LoLHR} for \ac{MORRDO} is given in Figure \ref{fig:FlowChartAdapt}.

Starting with some initial \ac{LHS} $\detm{X}^{0} \in \mathbb{R}^{m_0 \times n}$, the corresponding responses $\detm{Y}^{0} \in \mathbb{R}^{m_0 \times n_r}$ are computed and surrogate models are trained for each objective $f_i$ and limit state function $g_j$ using the data set $\mathbb{D}^0 = \{\detm{X}^{0}, \detm{Y}^{0}\}$. $m_0$ is the initial sample size and $n, n_r$ are the number of input dimensions and responses, respectively. Following, \ac{MORRDO} is conducted on the surrogate models and a set of input points $\detm{X}_{\mathbb{F}} \in \mathbb{R}^{m_{\mathbb{F}}, n}$ are recorded. $\detm{X}_{\mathbb{F}}$ contains the resulting Pareto frontier $\detm{X}^*_k$ as well as the prediction coordinates used for uncertainty quantification of solutions or designs in $\detm{X}^*_k$. These include the predictions used for the estimation of the moments $\operatorname{E}^{\boldsymbol{\theta}_{\stov{X}}}\left[f_k(\stov{X})\right],\operatorname{Var}^{\boldsymbol{\theta}_{\stov{X}}}\left[f_k(\stov{X})\right]$ and the probability of failure $P(\mathcal{F})$. 

\begin{figure}[H]
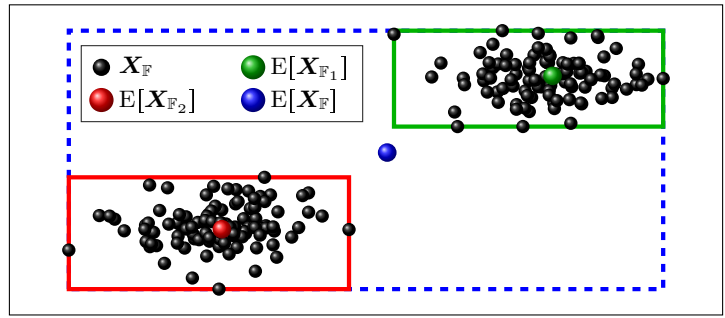

	\setlength\fheight{0.25\textwidth}%
	\setlength\fwidth{0.6\textwidth}%
	\centering
\ifusepdf%
	\includegraphics[]{Manuscript-figure\thepdffigno.pdf}%
	\refstepcounter{pdffigno}%
\else%
	\input{./Pictures/clusterbb/clusterbb.tex}%
\fi%

	\caption{$\detm{X}_{\mathbb{F}}$ with two distinct clusters. Analysing the clusters separately yields a more precise estimation of the bounding boxes and the mean values (green and red) compared to analysing all points at once (blue)} \label{fig:ClusterBB}
\end{figure}

Specifically, $\detm{X}_{\mathbb{F}}$ contains points from an orthogonal sampling following the joint distribution $F_{\stov{X}}(\cdot \; ; \boldsymbol{\theta}_{\stov{X}})$ as used for the robustness assessment as well as samples on the limit state function and in the region of failure $g_j(\detv{x}) \leq 0$.

\ac{UML} algorithm is used to find clusters in $\detm{X}_{\mathbb{F}}$ for an more accurate estimation of the local bounds (Figure \ref{fig:ClusterBB}) of disjoint Pareto optimal regions, as observed in multi modal functions (see Figure \ref{fig:ClusterBB}). Following, $\detm{X}^{k}  \in \mathbb{R}^{(m_0 + k m_s) \times n}$ is extended with new sampling points $\detm{X}^{k} \subset \detm{X}^{k+1}$ in these regions during the adaptation step. New points are obtained by generating and optimizing a \ac{LHS} within the estimated region regarding the minimum pairwise distance of the samples and the maximum correlation error (Eq. \ref{eq:AdaptObj}) of all sampling points. Optimization is conducted by switching the values among each column of the new points in $\detm{X}^{k+1}$ with a simulated annealing algorithm similar to the one proposed in \cite{Joseph2008}. 

The refined data set $\mathbb{D}^{k+1}$ is used for the training in the next iteration and the algorithm loops until some stopping criterion is met. In this work, only a sample budget is used as the termination criterion as this is often a limiting factor for practical applications. \ac{LoLHR} is run until the sample budget is depleted, \ie maximum number of samples are evaluated. The frontier $\detm{X}^*_{k=k_{\mathrm{max}}}$ as predicted by the final model trained on all data $\mathbb{D}^{k=k_{\mathrm{max}}}$ is used as the approximated solution of the optimization problem.

Depicted flow diagram is similar to other surrogate based sequential optimization algorithms except for the block denoted as adaptation. Also \ac{UML} is not included in many cases. The intuition here is to predict the region of interest at each step and acquire new samples based on these predictions. If the approximation is inaccurate, new samples are expected to improve the model approximation, such that different regions may be predicted in the subsequent steps. 
If the approximation is accurate, refinement continues to occur in the same region, where the local approximation error is expected to decrease, leading to a more precise solution. 

\subsection{Adaptation}
\begin{figure}[H]
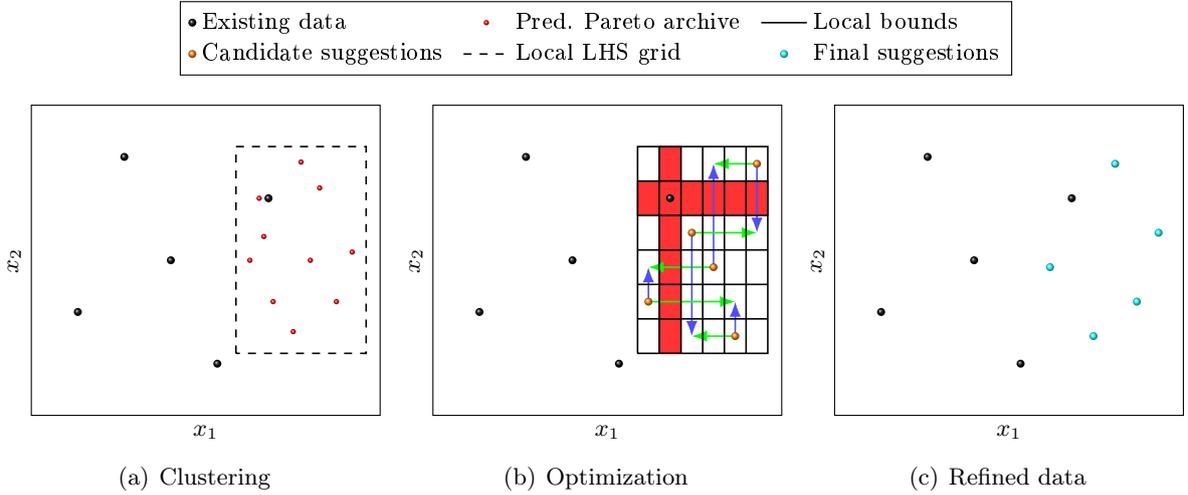
%
	\centering
	\subfigure{%
\ifusepdf%
	\includegraphics[]{Manuscript-figure\thepdffigno.pdf}%
	\refstepcounter{pdffigno}%
\else%
	\input{./Pictures/AdaptDoELegend.tex}%
\fi%
} \\
	\addtocounter{subfigure}{-1}
	\subfigure[Clustering]{
		\setlength\fheight{0.25\textwidth}%
		\setlength\fwidth{0.31\textwidth}%
\ifusepdf%
	\includegraphics[]{Manuscript-figure\thepdffigno.pdf}%
	\refstepcounter{pdffigno}%
\else%
	\input{./Pictures/AdaptDoE1.tex}%
\fi%

		\label{fig:Adapt1}}
	\subfigure[Optimization]{
		\setlength\fheight{0.25\textwidth}%
		\setlength\fwidth{0.31\textwidth}%
\ifusepdf%
	\includegraphics[]{Manuscript-figure\thepdffigno.pdf}%
	\refstepcounter{pdffigno}%
\else%
	\input{./Pictures/AdaptDoE2.tex}%
\fi%

		\label{fig:Adapt2}}
	\subfigure[Refined data]{
		\setlength\fheight{0.25\textwidth}%
		\setlength\fwidth{0.31\textwidth}%
\ifusepdf%
	\includegraphics[]{Manuscript-figure\thepdffigno.pdf}%
	\refstepcounter{pdffigno}%
\else%
	\input{./Pictures/AdaptDoE3.tex}%
\fi%

		\label{fig:Adapt3}}
	\caption{Visualization of the adaptation step}
	\label{fig:Adaptation}
\end{figure}

Adaptation is illustrated in Figure \ref{fig:Adaptation}. Bounds are obtained first using the clustered points $\detm{X}^s_{\mathbb{F}} \subseteq \detm{X}_{\mathbb{F}}$ as 
\changemine{\begin{equation}
\begin{gathered}
\detv{x}^l_{k, s} = \min(\detm{X}^s_{\mathbb{F}}) \\
\detv{x}^u_{k, s} = \max(\detm{X}^s_{\mathbb{F}})
\end{gathered} \label{eq:boundchoicebase}
\end{equation}}
where $\min(\cdot)$ and $\max(\cdot)$ are computed along each input dimension.

\changemine{This procedure may result in too small bounds, leading to early convergence and decreasing the global exploration behaviour, for example due to low number of Pareto solutions. As a precaution, a minimum size $\Delta\detv{x}_{\mathrm{min},s}$ is used to bound the size of the clusters on the lower end, defined as a length in each dimension. First, the bin size $\detv{b}_{k+1}$ for the \ac{LHS} including the new points is computed
\begin{equation}
\detv{b}_{k+1} = \frac{\detv{x}^u - \detv{x}^l}{m_{k+1}}
\end{equation}
and used to compute the size of $m_s$ bins, where $m_s$ is the number of new samples to be adapted in the cluster $s$. Following, the minimum size is computed as 
\begin{equation}
\Delta\detv{x}_{\mathrm{min},s} = \detv{b}_{k+1} m_s
\end{equation}
Finally, the bounds are set as:
\begin{equation}
\begin{gathered}
\detv{x}^l_{k, s} = \textcolor{red}{\min}\left(\min(\detm{X}^s_{\mathbb{F}}), \boldsymbol{\mu}_{\detm{X}^s_{\mathbb{F}}} - 0.5 \Delta\detv{x}_{\mathrm{min},s}\right) \\
\detv{x}^u_{k, s} = \textcolor{red}{\max}\left(\max(\detm{X}^s_{\mathbb{F}}), \boldsymbol{\mu}_{\detm{X}^s_{\mathbb{F}}} + 0.5 \Delta\detv{x}_{\mathrm{min},s}\right)
\end{gathered} \label{eq:boundchoice}
\end{equation}
}

Following, an \ac{LHS} grid is placed within the local bounds $\detv{x}^l_{k, s}, \detv{x}^u_{k, s}$ (Figure \ref{fig:Adapt1}). The number of bins in the local \ac{LHS} are increased until $m_s$ empty bins occur for at least one input dimension $x_i$. Another choice would be to increase the bins until all dimensions have at least $m_s$ empty bins but this increases the sparsity of the local \ac{LHS} and thus diminishes its space filling properties. The candidate for the local \ac{LHS} is thus generated (orange points in Figure \ref{fig:Adapt2}) and optimized using simulated annealing depicted as arrows. Resulting sampling points in Figure \ref{fig:Adapt3} are used to extend the data set for the next iteration.

A similar method using \ac{LHS} schemes and sequential sampling was proposed in \cite{Wang2003} for deterministic optimization in combination with polynomial regression. However, already existing points are ignored in \cite{Wang2003} and a small \ac{LHS} is mapped to a larger one, preventing the calculation of any metrics like the minimum distance or correlation. Furthermore, optimization of the \ac{LHS} or clustering is not conducted. Here, the following objective function is used for the optimization
\begin{equation}
\begin{gathered}
f_M(\detm{X}^k) = f_D(\detm{X}^k)  +  f_{\rho}(\detm{X}^k) \\
f_D(\detm{X}^k) = \mathrm{log}(d_{\mathrm{max}}) - \mathrm{log}(\mathrm{min}\left(\mathrm{pdist}(\detm{X}^k)\right))  \\
f_{\rho}(\detm{X}^{k,L}) = \mathrm{log}\left(\mathrm{max}\left( \left| \boldsymbol{\rho}_{\mathbb{F} \mathbb{F}} - \boldsymbol{\rho}_{\detm{X}^{k, L} \detm{X}^{k, L}} \right| \right) \right) 
\end{gathered}\label{eq:AdaptObj}
\end{equation}
where $\mathrm{pdist}(\detm{X}^k)$ denotes the pairwise distances of $\detm{X}^k$ and $\boldsymbol{\rho}_{\mathbb{F} \mathbb{F}}$ is the Pearson correlation coefficient of points $\detm{X}^s_{\mathbb{F}}$ in the current cluster. $\boldsymbol{\rho}_{\detm{X}^{k, L} \detm{X}^{k, L}}$ is the correlation of in $\detm{X}^k$, that lie within the local bounds $\detv{x}^l_{k,L}$ and $\detv{x}^u_{k,L}$.  

$f_M$ consists of two terms. $f_D$ represents the minimum pairwise distance which is maximized if $f_D$ is minimized. Minimum pairwise distance is often maximized to improve the space filling criterion \citepeg{Joseph2008}. The second term $f_{\rho}$ is an error metric the difference $|\boldsymbol{\rho}_{\mathbb{F} \mathbb{F}} - \boldsymbol{\rho}_{\detm{X}^{k, L} \detm{X}^{k, L}}|$. For stationary sampling, global correlation error is used instead, \ie the local and global bounds are the same and $\boldsymbol{\rho}_{\mathbb{F} \mathbb{F}}=\detm{I}_{d}$ where $\detm{I}_{d}$ is the $d \times d$ identity matrix. Using global correlation error for stationary sampling is also common \citep{Iman1982} to achieve a more space filling sample set. However, using the local correlation error is not as common but, it makes sense within the proposed framework. \ac{LoLHR} seeks to improve accuracy in the region of interest, which an arbitrarily shaped subset of the global space. Local correlation error wrt. $\detm{X}^s_{\mathbb{F}}$ is minimized to increase the density of the samples around this region.

\begin{figure}[H]
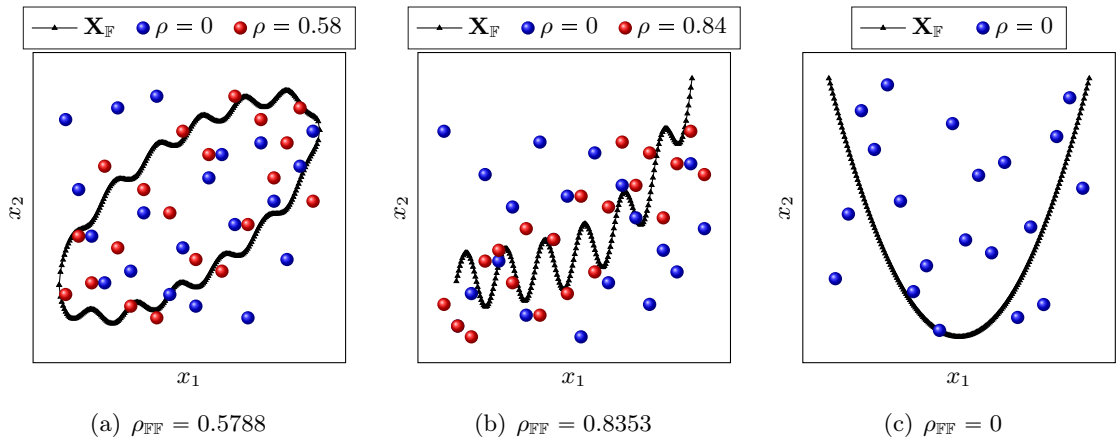
%
 	\centering
 	\subfigure[$\rho_{\mathbb{F}\mathbb{F}} = 0.5788$]{
 		\setlength\fheight{0.25\textwidth}%
 		\setlength\fwidth{0.25\textwidth}%
 		%
\ifusepdf%
	\includegraphics[]{Manuscript-figure\thepdffigno.pdf}%
	\refstepcounter{pdffigno}%
\else%
	\input{./Pictures/correlatedlsf/ellipsis.tex}%
\fi%

 		\label{fig:corrlsf_ellipsis}
 	} 
 	\subfigure[$\rho_{\mathbb{F}\mathbb{F}}= 0.8353$]{
 		\setlength\fheight{0.25\textwidth}%
 		\setlength\fwidth{0.25\textwidth}%
 		%
\ifusepdf%
	\includegraphics[]{Manuscript-figure\thepdffigno.pdf}%
	\refstepcounter{pdffigno}%
\else%
	\input{./Pictures/correlatedlsf/expsin.tex}%
\fi%

 		\label{fig:corrlsf_expsin}
 	}
 	\subfigure[$\rho_{\mathbb{F}\mathbb{F}} = 0$]{
 		\setlength\fheight{0.25\textwidth}%
 		\setlength\fwidth{0.25\textwidth}%
 		%
\ifusepdf%
	\includegraphics[]{Manuscript-figure\thepdffigno.pdf}%
	\refstepcounter{pdffigno}%
\else%
	\input{./Pictures/correlatedlsf/quad.tex}%
\fi%
 
 		\label{fig:corrlsf_quad}
 	}
 	\caption{Visualization of example points $\detm{X}_{\mathbb{F}}$ with non-linear dependence as well as samples $\detm{X}^k$ with matching and $0$ correlation.}
 	\label{fig:corrlsf}
 \end{figure}

Consider the regions of interest in Figures \ref{fig:corrlsf_ellipsis} and \ref{fig:corrlsf_expsin}. In these cases, placing the new points in $\detm{X}^k$ with the same correlation as $\detm{X}_{\mathbb{F}}$ increases the density of the samples around the region of interest. However in Figure \ref{fig:corrlsf_quad}, where the relationship is quadratic, the Pearson correlation coefficient is $0$ as this dependency measure cannot detect non-monotonic relationships. Thus, $f_{\rho}$ is essentially used to quickly estimate a bandwidth for the new points, such that the sampling density near the region of interest increases. 

\section{Toy examples}\label{sec:results}
Results are discussed in the following for three test cases and compared to some of the already existing methods. Since the solutions are not points but rather a set of points, hypervolume indicator (HVI) was used to evaluate statistics of the results over multiple runs. A description of (HVI) can be found in the \ref{app:opti}. 

As the baseline for comparisons, two direct strategies using \ac{NSGA}-II was tested. For the longer version, $100$ generations with a population size of $100$. A shorter version with $10$ generations and a population size of $10$ was also ran for closing the gap between the surrogate based strategies regarding the number of samples. In both cases, direct strategies required magnitudes of order more samples compared to the surrogate based strategies. \changerev{Moreover, results with random sampling are also given for comparison. In this case, $m$ randomly chosen designs were validated where $m$ is the number of function evaluations used for surrogate based methods.}

When using surrogate based strategies with \ac{NSGA}-II, a population size of $200$ was chosen instead using $200$ generations since these evaluations are assumed to be much cheaper than the evaluation of the true numerical model. Moreover, stationary sampling, \ie data generated in one-shot, was also used as a baseline, on which the proposed sequential sampling strategies should improve.

\subsection{Simple analytical 2-d example}\label{sec:ex1}
The proposed method was tested first on a simple problem with two input parameters, two objectives and one limit state:
\begin{equation}
\begin{gathered}
f_1 (x_1,x_2) =  \frac{1}{7}(5 \sqrt{2} - x_1 - x_2)   \\
f_2 (x_1,x_2) =\frac{1}{180} \sum_{i=1}^{2}(x_i^4 - 16x_i^2 + 5x_i) \\
g(x_1,x_2) = \left(\frac{x_1^2 + x_2}{1.81} - 11\right)^2 + \left(\frac{x_1 + x_2^2}{1.81} - 7\right)^2 - 45 \\
\end{gathered}
\end{equation}

The \ac{RRDO} problem is formulated as follows
\begin{equation}
\begin{gathered}
\argmin_{\boldsymbol{\mu}_{\stov{X}}} \; \mathrm{E}^{\boldsymbol{\mu}_{\stov{X}}}\left[f_k(\detv{x})\right] + 1.96 \mathrm{Var}^{\boldsymbol{\mu}_{\stov{X}}}(f_k(\detv{x})) \qquad  k \in \; \{1,2\}  \\
\textit{\footnotesize s.t.} \quad P(\mathcal{F})  \leq 10^{-6}  \\
X_1 \sim \mathcal{N}\left(\mu_{X_1}, 0.2^2\right) \quad X_2 \sim \mathcal{N}\left(\mu_{X_2}, 0.2^2\right) \quad -5 \leq \mu_{X_i} \leq 5
\end{gathered}\label{eq:RRDOProblem1}
\end{equation}
Input variables $\stov{X}$ are assumed to be independent and $\mathcal{N}(\mu, \sigma^2)$ denotes a normal distribution with mean $\mu$ and variance $\sigma^2$. Boundaries $\detv{x}^{l,D}$ and $\detv{x}^{u,D}$ of the initial \ac{LHS} and the stationary sampling were chosen as the one sided $\alpha = 99.9 \%$ confidence bound of each marginal distribution at the optimization boundaries $\detv{x}^l=[-5, -5]$ and $\detv{x}^u=[5, 5]$
\begin{equation}
x^{l,D}_i =  F^{-1}_{X_i} \left( 1 - \alpha ; \mu_{X_i} = x^l_i\right) \qquad 
x^{u,D}_i =  F^{-1}_{X_i} \left( \alpha ; \mu_{X_i} = x^u_i\right)
\end{equation}
where $F^{-1}_{X_i}(\cdot)$ is the inverse of the marginal CDF of $i$-th input variable $X_i$.


Since the uncertainty assessment is directly coupled with the optimizer, every single deterministically feasible design \ie candidate solution requires both a robustness and a reliability assessment. For the robustness assessment, an orthogonal sampling with $200$ samples and the corresponding joint distribution was used throughout this work. 

In this example, \ac{DS} was used with $160$ directions, $20$ initial points in each direction to find the bracketing interval for the root finding algorithm. For deterministically feasible designs, this required a total of $3781.45$ points on average in the interval $[3201, 4227]$ for the estimation of $P(\mathcal{F})$. Long direct strategy required $10000$ design evaluations for \ac{NSGA}-II. On average $8832.4$ of these designs required an uncertainty analysis. Thus, a total of $3.38 \cdot 10^{7}$ samples were required for these results on average. $68.3$ deterministically feasible designs were found on average using the short strategy which resulted in $2.58 \cdot 10^{5}$ samples.

\begin{figure}[ht]
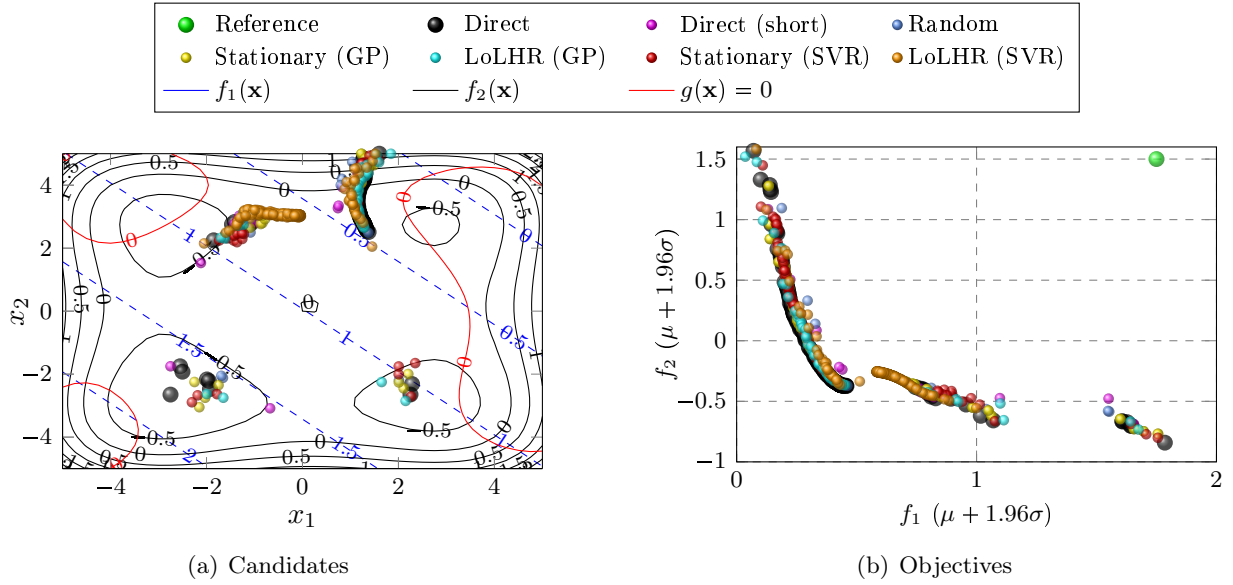

	\centering
	\setlength\fheight{0.35\textwidth} 
	\setlength\fwidth{0.48\textwidth}
	\subfigure{%
\ifusepdf%
	\includegraphics[]{Manuscript-figure\thepdffigno.pdf}%
	\refstepcounter{pdffigno}%
\else%
	\input{./Pictures/Adaptivity/AnalyticNew/example1/Legend}%
\fi%
} \\
	\addtocounter{subfigure}{-1}
	\subfigure[Candidates \label{fig:ParetoAnalMLContour}]{
\ifusepdf%
	\includegraphics[]{Manuscript-figure\thepdffigno.pdf}%
	\refstepcounter{pdffigno}%
\else%
	\input{./Pictures/Adaptivity/AnalyticNew/example1/ParetoArchive.tex}%
\fi%

	}\hfill 
	\subfigure[Objectives \label{fig:ParetoAnalMLPareto}]{
\ifusepdf%
	\includegraphics[]{Manuscript-figure\thepdffigno.pdf}%
	\refstepcounter{pdffigno}%
\else%
	\input{./Pictures/Adaptivity/AnalyticNew/example1/ParetoFrontValid.tex}%
\fi%

	}\caption{Validated Pareto frontiers for the first example after $10$ runs. Direct refers to the surrogate-free optimization with $100$ generations and the populations size of $100$, whereas Direct (short) refers to a run with $10\times10$ design evaluations. Frontier with the highest \ac{HVI} is shown for each method.}
	\label{fig:ParetoAnalML}
\end{figure}

In contrast, $64$ points were used with \ac{GP} and \ac{SVR} for surrogate based methods, several orders of magnitude less than the direct methods. Nevertheless, the resulting frontiers were compared for the evaluation of performance loss due to surrogate modelling. For the LoLHR, $m_0=32$ initial samples were generated and $4$ steps each with $m_s = 8$ samples were taken. $10$ runs were conducted with each strategy and the best Pareto frontier \ie the one with the highest \ac{HVI} is depicted in Figure \ref{fig:ParetoAnalML}. The number of iterations and the population size were increased to $200$ when optimizing the surrogate models, as this did not increase the number of samples computed on the true function.



\begin{table}%
	\centering
	\footnotesize
	\caption{Results of the first example of each strategy after $10$ runs. Bold numbers represent the best results while \changefinal{the best among the surrogate based solutions are underlined}. $\tilde{\mu}_h$, $\tilde{\sigma}_h$, $\min_h$, $\max_h$ represent the sample average, sample standard deviation, minimum and maximum hypervolume indicator in this order. Furthermore, the average number of unreliable designs $\tilde{\mu}_\mathcal{F}$, Pareto solutions $\tilde{\mu}_p$ and function evaluations $\tilde{\mu}_m$ are also given.}
	\label{tab:ParetoAnalErrors}
	\bgroup
	\def\arraystretch{1.1}
	\begin{tabular}{l | c c c c c c c}
		Method & $\tilde{\mu}_h$ & $\tilde{\sigma}_h$ & $\min_h$ & $\max_h$ & $\tilde{\mu}_\mathcal{F}$ & $\tilde{\mu}_p$ & $\tilde{\mu}_m$\\
		\hline
		\changerev{Random} & $2.289$ & $0.482$ & $1.697$ & $2.842$ & $33.2$ & $7.0$ & $2.60\cdot 10^5$ \\
		Direct & $\bm{3.059}$ & $\bm{0.028}$ & $\bm{3.010}$ & $\bm{3.104}$  & $\bm{0}$ & $\bm{128.8}$ & $3.38 \cdot 10^{7}$\\
		Direct (short) & $2.572$ & $0.185$ & $2.355$ & $2.889$ & $\bm{0}$ & $20.9$ & $2.58 \cdot 10^{5}$\\
		\ac{GP} (stat.) &  $2.532$ & $0.517$ & $1.775$ & $3.072$ & $38.3$ & $34.6$ & $\underline{\bm{64}}$\\
		\ac{GP} (LoLHR) & $2.831$ & $0.371$ & $1.860$ & $\underline{3.094}$ &  $33.1$ & $51.5$ & $\underline{\bm{64}}$\\
		\ac{SVR} (stat.) &  $2.628$ & $0.368$ & $1.739$ & $3.042$ & $45.8$ & $68.5$ & $\underline{\bm{64}}$\\
		\ac{SVR} (LoLHR) & $\underline{2.863}$ & $\underline{0.279}$ & $\underline{2.118}$ & $3.086$ &  $\underline{29.2}$ & $\underline{103.3}$ & $\underline{\bm{64}}$\\
	\end{tabular}
	\egroup
\end{table}

The reference point for the \ac{HVI} calculation was chosen as $[1.75, 1.5]$. This point was chosen after investigating the results of the direct strategy. For practical problems, it is often possible to choose a more meaningful reference point before the optimization according to the domain specific requirements of the application. Reported number of samples for the surrogate based strategies does not include the validation, which also has a high computational burden. Although every Pareto point is validated in the following, often the validation of a few samples are sufficient for finding a preferred solution in real world applications, if a preference regarding the trade-off is present. Nonetheless, the validation of the estimated failure probability $P(\mathcal{F})$ is done using \ac{DS} for all estimated Pareto designs. In contrast, direct methods do not require a validation as the true models are used throughout the optimization. \changerev{However, a validation of the random sampling has to be conducted to be able to differentiate the reliable and Pareto optimal samples from the rest.}

It can be seen that in the best case, surrogate based strategies acquire very similar results to the best case of the direct optimization, despite the large difference in the required number of samples (Figure \ref{fig:ParetoAnalML}). Nonetheless, the long direct approach achieved on average better results with a lower variance compared to the surrogate based strategies. Furthermore, both for the stationary sampling and \ac{LoLHR}, \ac{SVR} acquired on average slightly better results in this example although the best \ac{LoLHR} run with \ac{GP} achieved better results than \ac{SVR}. 

Detailed results are given in Table \ref{tab:ParetoAnalErrors} for $10$ runs, where $\tilde{\mu}_h$ and $\tilde{\sigma}_h$ correspond to the sample average and the standard deviation of the \ac{HVI}, as calculated from the acquired frontiers. Furthermore for \ac{SML} based strategies, some designs in the predicted frontier may not be feasible upon validation. Thus, Table \ref{tab:ParetoAnalErrors} includes the average number of infeasible designs $\tilde{\mu}_\mathcal{F}$ as well as the average number of predicted Pareto optimal designs $\tilde{\mu}_p$ \changemine{excluding the infeasible and unreliable ones.}

\subsection{Tricky analytical 2-d example} \label{sec:ex2}
The following functions were chosen as a multi modal \ac{MORRDO} example
\begin{equation}
\begin{aligned}
f_1 (\detv{x}) &= \frac{(x_1^4 + x_2^4) - 16 (x_1^2 + x_2^2) + 5(x_1 + x_2)}{180}\\
f_2 (\detv{x}) &= \frac{\left(x_1 -2.25\right)^2 + \left(x_2 -2.25\right)^2}{50} \\
g(\detv{x}) &= 7 - \sum_i^2 \left( \frac{x_i}{1.475} \right)^2 - 5 \cos \left(2 \pi \frac{x_i}{1.475} \right)
\end{aligned}\label{eq:RRDOProblem2}
\end{equation}
Furthermore, the \ac{RRDO} problem was defined as
\begin{equation}
\begin{gathered}
\argmin_{\boldsymbol{\mu}_{\stov{X}}} \; \mathrm{E}^{\boldsymbol{\mu}_{\stov{X}}}\left[f_k\right] + 1.96 \mathrm{Var}^{\boldsymbol{\mu}_{\stov{X}}}(f_k) \;  k \in \{1,2\}  \\
\textit{\footnotesize s.t.} \quad P(\mathcal{F})  \leq 10^{-2}  \\
X_1 \sim \mathcal{N}\left(\mu_{X_1},0.15^2\right) \qquad X_2 \sim \mathcal{U}\left(\mu_{X_2},\frac{0.5^2}{12}\right) \qquad -4.5 \leq \mu_{X_i} \leq 4.5
\end{gathered}
\end{equation}
where $\mathcal{U}(\mu,\sigma^2)$ denotes a uniform distribution with the mean $\mu$ and variance $\sigma^2$ instead of the conventional notation using the lower and upper bounds. Notice that even though $2$ dimensional, the limit state function in this example as taken from \citep{Echard2011} is not easy to approximate with \ac{SML} methods.  The total number of samples was set to $128$ for \ac{SML} based methods because of this difficulty. Furthermore, \ac{MC} with $10^6$ was used for reliability analysis. 

\changerev{The sequential sampling method proposed by \cite{Gu2013, Gu2014} for solving \ac{MORRDO} problems in combination with both \ac{GP} and \ac{SVR} was implemented and tested for comparison. Their method consists of sampling the \textit{elbow point} as well as other randomly selected designs from the Pareto frontier. Elbow point is defined in this context as the point closest to the origin on the unit scaled Pareto frontier space, where each objective is scaled to the interval $[0, 1]$.}

For \ac{LoLHR} and \cite{Gu2013}, $m_0 = 64$ initial samples were refined in $4$ steps. For each step, $16$ new samples are generated to acquire the displayed Pareto frontier. Results in the best case and on average were slightly better compared to the direct approach, probably since more iterations for the optimization on the surrogate models was allowed.


\begin{figure*}
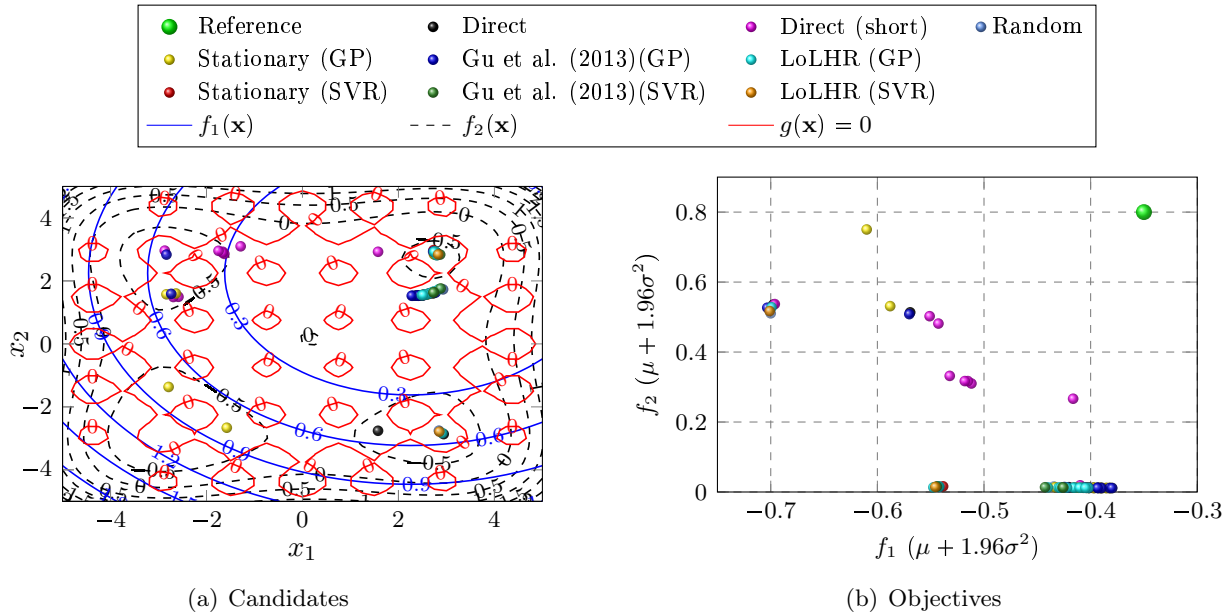

	\centering
	\setlength\fheight{0.35\textwidth} 
	\setlength\fwidth{0.48\textwidth}
	\subfigure{%
\ifusepdf%
	\includegraphics[]{Manuscript-figure\thepdffigno.pdf}%
	\refstepcounter{pdffigno}%
\else%
	\input{./Pictures/Adaptivity/AnalyticNew/example2/Legend}%
\fi%
} \\
	\addtocounter{subfigure}{-1}
	\subfigure[Candidates \label{fig:ParetoAnalCosMLContour}]{
\ifusepdf%
	\includegraphics[]{Manuscript-figure\thepdffigno.pdf}%
	\refstepcounter{pdffigno}%
\else%
	\input{./Pictures/Adaptivity/AnalyticNew/example2/ParetoArchive.tex}%
\fi%
 \label{fig:ParetoArchiveCos}
	}\hfill 
	\subfigure[Objectives \label{fig:ParetoAnalCosMLPareto}]{
\ifusepdf%
	\includegraphics[]{Manuscript-figure\thepdffigno.pdf}%
	\refstepcounter{pdffigno}%
\else%
	\input{./Pictures/Adaptivity/AnalyticNew/example2/ParetoFrontValid.tex}%
\fi%
 \label{fig:ParetoFrontCos}
	}\caption{Results of the second example after $10$ runs. Frontier with the highest \ac{HVI} is shown for each method.}
	\label{fig:ParetoAnalCosML}
\end{figure*}

Figure \ref{fig:ParetoAnalCosML} depicts the resulting Pareto frontiers with the highest HVI for each method. Stationary sampling and the short direct approach could only find a limited part of the Pareto frontier even in the best case in contrast to \ac{LoLHR} and the long direct approaches. The same holds for \cite{Gu2013} with \ac{SVR} as proposed in \cite{Gu2014}. The missing parts were the leftmost and bottom-most portions of the frontier in Figure \ref{fig:ParetoFrontCos} as well as the top and bottom right corners of the feasible region in Figure \ref{fig:ParetoArchiveCos}. For the surrogate based strategies, the models inaccurately predicted one or both of those regions to be unreliable. This was not the case for \ac{LoLHR} based strategies as well as \cite{Gu2013} in combination with \ac{GP}. 

\changerev{Interestingly, random sampling managed to find both of these in the best case but missed them in all other cases. Nonetheless, random sampling delivered on average $106.9$ unreliable designs ($101$ in best case) out of $128$ and only $4.6$ Pareto optimal ones ($9$ in the best case). In other words, $< 16.5 \%$ of the random designs were reliable on average and $< 3.6 \%$ of them were Pareto optimal with respect to a set of other $127$ random designs from the same run.}



Detailed results of all runs are given in Table \ref{tab:ParetoAnalCosErrors}. The reference point was chosen as $[-0.35, 0.8]$ for the \ac{HVI} calculation. Both \ac{GP} based stationary and \ac{LoLHR} strategies acquired a lower variance of \ac{HVI} compared to other strategies including the direct ones, again probably due to the higher number of allowed iterations on the surrogate models. \ac{LoLHR} with \ac{GP} achieved on average the best results in this example, although the results in the best case were slightly worse compared to \ac{LoLHR} with \ac{SVR}. All investigated \ac{HVI} metrics were better for \ac{LoLHR} strategies compared to other surrogate based strategies. Furthermore, the same holds for \ac{LoLHR} with \ac{GP} compared to the direct approaches.

\begin{table*}%
	\centering
	\footnotesize
	\caption{\ac{HVI} statistics of each strategy after $10$ runs. Bold numbers represent the best results while \changefinal{the best among the surrogate based solutions are underlined}. $\tilde{\mu}_h$, $\tilde{\sigma}_h$, $\min_h$, $\max_h$ are same as Table \ref{tab:ParetoAnalErrors}.}
	\label{tab:ParetoAnalCosErrors}
	\bgroup
	\def\arraystretch{1.1}
	\begin{tabular}{l | c c c c c c c}
		Method & $\tilde{\mu}_h$ & $\tilde{\sigma}_h$ & $\min_h$ & $\max_h$ & $\tilde{\mu}_\mathcal{F}$ & $\tilde{\mu}_p$ & $\tilde{\mu}_m$\\
		\hline
		\changerev{Random} & $0.0539$ & $0.0582$ & $-0.0905$ & $0.1432$ & $106.9$ & $4.6$ & $1.28\cdot 10^8$ \\
		Direct & $0.1562$ & $0.0300$ & $0.1004$ & $0.1936$  & $\bm{0}$ & $25.4$ & $3.42 \cdot 10^{10}$\\
		Direct (short) & $0.0745$ & $0.0470$ & $-0.0405$ & $0.1505$ & $\bm{0}$ & $12.0$ & $2.83 \cdot 10^{8}$\\
		\ac{GP} (stat.) &  $0.1418$ & $0.0232$ & $0.0985$ & $0.1650$ & $\underline{5.5}$ & $23.9$ & $\underline{\bm{128}}$\\
		\ac{GP} (LoLHR) & $\underline{\bm{0.1644}}$ & $\underline{\bm{0.0226}}$ & $\underline{\bm{0.1232}}$ & $0.1967$ &  $6.4$ & $\underline{\bm{41.3}}$ & $\underline{\bm{128}}$\\
		\changerev{\ac{GP} \citep{Gu2013}} & $0.1540$ & $0.0309$ & $0.1023$ & $0.1956$ &  $6.1$ & $27.9$ & $\underline{\bm{128}}$\\
		\ac{SVR} (stat.) &  $0.0642$ & $0.0563$ & $0.0$ & $0.1476$ & $28.6$ & $30.1$ & $\underline{\bm{128}}$\\
		\ac{SVR} (LoLHR) & $0.1252$ & $0.0464$ & $0.1085$ & $\underline{\bm{0.1976}}$ & $9.7$ & $12.1$ & $\underline{\bm{128}}$\\
		\changerev{\ac{SVR} \citep{Gu2013}} & $0.0807$ & $0.0337$ & $0.0211$ & $0.1525$ & $26.2$ & $30.5$ & $\underline{\bm{128}}$\\
	\end{tabular}
	\egroup
\end{table*}

Although the method proposed by \cite{Gu2013} improved the results compared to stationary sampling, the difference was smaller compared to \ac{LoLHR}.  Nonetheless, all \ac{SML} based strategies also predicted some unreliable designs (\ie $P(\mathcal{F}) > 10^{-2}$) to be Pareto optimal. This is generally a big disadvantage of using surrogate based methods for strongly constrained problems. 


\subsection{Short column under oblique bending} \label{sec:ex3}

Mass of a short column with rectangular cross-section subject to biaxial bending and tensile stress is to be minimized in this example, while keeping the $P(\mathcal{F})$ smaller than a target level $1.35 \cdot 10^{-3}$. Buckling is ignored due to the assumption of shortness. 
Failure is defined as the ultimate stress $S$ exceeding the material yield stress $R$. Similar to \cite{Royset2011,Dubourg2011,Dubourg2011a}, the limit state function is given as
\begin{equation}
g\left(\stov{X}\right) = 1 - \frac{4 M_1}{BH^2R} - \frac{4 M_2}{B^2HR} - \left(\frac{F_{\mathrm{ax}}}{BHR}\right)^2 \label{eq:mechexlsf}
\end{equation}
For $g < 0$, the ultimate stress is larger than the yield stress $R$, causing plastic deformations that damage the structure and render the design infeasible. The derivation of Eq. \ref{eq:mechexlsf} is available in \cite{Dubourg2011a} and omitted here. In general, reducing the cross-section area reduces the mass but increases the ultimate stress. The original problem thus aims to reduce the mass as much as the yield stress of the material and the parameter uncertainties allow.

\changerev{Note that this example with different parameter distributions was investigated by \cite{Aoues2010,Lobato2020}. Especially the strategy in \cite{Lobato2020} is interesting due to their \ac{MORRDO} formulation. However, a comparison to a surrogate based method was preferred here. Moreover, since \cite{Dubourg2011a} also reports results for other methods described in \cite{Aoues2010}, the formulation below was chosen and converted to a \ac{MORRDO} problem similar to \cite{Lobato2020}, where $>3 \cdot 10^6$ function evaluations were required for finding the Pareto frontier.}

The design parameters cross-section width $B$ and height $H$ in the original problem \citep{Royset2011} are deterministic but small variances were assigned to $B$ and $H$ by \cite{Dubourg2011,Dubourg2011a}, defining the \ac{RBDO} problem as:
\begin{equation}
\begin{gathered}
\argmin_{\mu_B, \mu_H} \; \mu_B \mu_H(1 + 100 P(\mathcal{F}))\\
\textit{\footnotesize s.t.} \; P(\mathcal{F}) \leq 1.35 \cdot 10^{-3} \\
0.5 \leq \frac{\mu_B}{\mu_H} \leq 2 \\
100 \leq \mu_B, \mu_H \leq 1000
\end{gathered}\label{eq:ex4orig}
\end{equation}

The distribution of the model parameters are given in Table \ref{tab:ex3dists}. Notice that the failure probability $P(\mathcal{F})$ also contributes to the objective value in contrast to the previous example. The objective function is a weighted combination of two contradictory or competing terms; the smaller values of the design variables $B, H$ reduce the material costs but lead to a higher stress, increasing $P(\mathcal{F})$. 

If the cost of failure or the relative importances of the objectives are not known, deciding for the most appropriate weights may not be trivial in practical applications, especially without a prior observation of the trade-off between the competing objectives. Therefore, we propose to solve the corresponding \ac{MORBDO} or the \ac{MORRDO} problem directly. Not only does this approach captures the trade-off behaviour, but it also gives a good approximation of the SO solutions of multiple objective weight choices with a similar number of samples (see Table \ref{tab:ex3dists}).  
Hence, the objective in Eq. \ref{eq:ex4orig} is converted to the following two objectives first of which ($f_1$) is deterministic for comparability with the original work:
\begin{equation}
\begin{gathered}
\argmin_{\mu_B, \mu_H} \; f_1, f_2 \\
f_1(\mu_B, \mu_H) = \mu_B \cdot \mu_H \\
f_2\left(\stov{X}\right) = P(\mathcal{F}) = P(g\left(\stov{X}\right) < 0) \\
\stov{X} = [M_1, M_2, F_{\mathrm{ax}}, R, B, H]
\end{gathered}\label{eq:ex4mo}
\end{equation}

Furthermore, since $P(\mathcal{F})$ is used as an objective to minimize the probability that the ultimate stress becomes larger than the yield stress under uncertainty, the criterion regarding the failure probability in Eq. \ref{eq:ex4orig} is removed. The constraints and bounds are kept as before. The minimum estimated $P(\mathcal{F})$ is limited to $1.35 \cdot 10^{-5}$. This reflects the expected precision of \ac{DS} used with the preferred settings, since the maximum search radius for the failure region in each direction was chosen as described in Eq. \ref{eq:Dsrmax}. Hence, the actual definition of the second objective function $f_2$ is given as
\begin{equation}
f_2\left(\stov{X}\right) = \max\left(1.35 \cdot 10^{-5}, P(\mathcal{F})\right)\\
\end{equation}

\begin{figure}[ht]
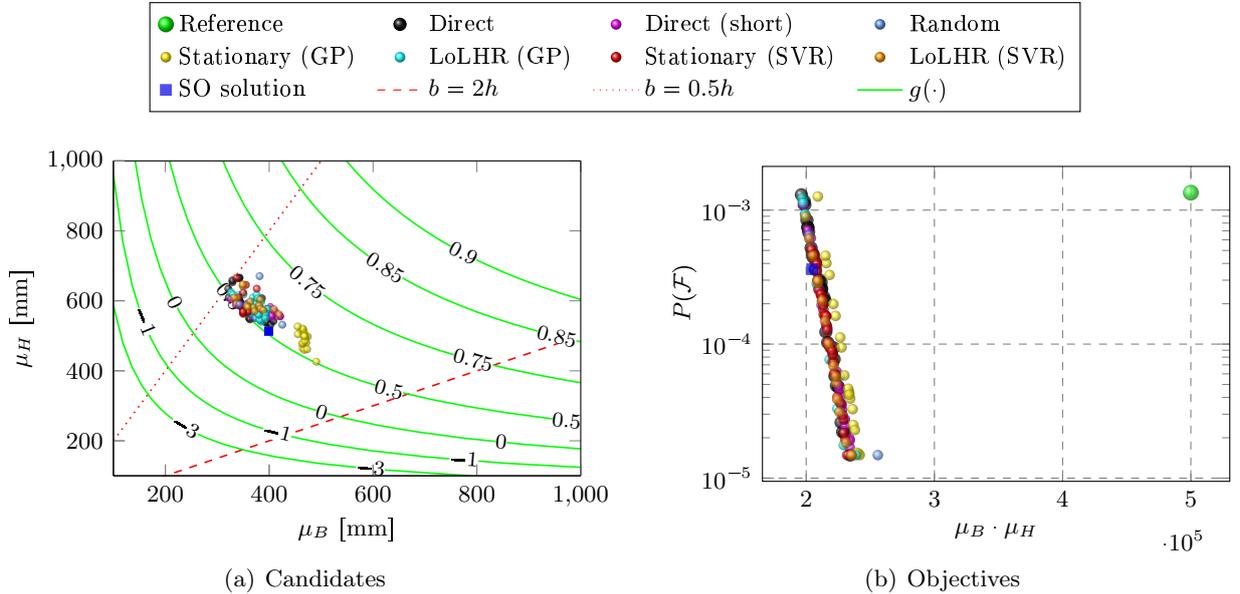

	\centering
	\setlength\fheight{0.35\textwidth} 
	\setlength\fwidth{0.47\textwidth}
	\subfigure{%
\ifusepdf%
	\includegraphics[]{Manuscript-figure\thepdffigno.pdf}%
	\refstepcounter{pdffigno}%
\else%
	\input{./Pictures/Adaptivity/AnalyticNew/example4/Legend}%
\fi%
} \\
	\addtocounter{subfigure}{-1}
	\subfigure[Candidates \label{fig:ParetoArchiveMech}]{
\ifusepdf%
	\includegraphics[]{Manuscript-figure\thepdffigno.pdf}%
	\refstepcounter{pdffigno}%
\else%
	\input{./Pictures/Adaptivity/AnalyticNew/example4/ParetoArchive.tex}%
\fi%
 
	}\hfill
	\subfigure[Objectives \label{fig:ParetoFrontMech}]{
\ifusepdf%
	\includegraphics[]{Manuscript-figure\thepdffigno.pdf}%
	\refstepcounter{pdffigno}%
\else%
	\input{./Pictures/Adaptivity/AnalyticNew/example4/ParetoFrontValid.tex}%
\fi%

	}\caption{Results of the mechanical example after $10$ runs. $c_g(\cdot)$ represents 2-d section of the limit state function (Eq. \ref{eq:mechexlsf}), where all but the optimization variables were set to their mean values as defined by their marginal distributions. Frontier with the highest \ac{HVI} among all runs is shown for each method. \ac{SO} solution is taken from \citep{Dubourg2011}.}
	\label{fig:ParetoMech}
\end{figure}

Results of the best runs are depicted in Figure \ref{fig:ParetoMech}. $128$ samples were used for obtaining the surrogate based solutions. For \ac{LoLHR}, $m_0=64$ start samples were generated randomly and four steps with $m_s=16$ were conducted. Furthermore, the best solution \changerev{(RIA in Table \ref{tab:ParetoMechErrors})} of the \ac{SO} problem from \citep{Dubourg2011} is shown for comparison. Authors report that this solution required $9472$ calculations without using a surrogate model but they also obtain a surrogate based solution using $140$ samples, which is visually indistinguishable from the displayed solution \changerev{(GP (NRBDO) in Table \ref{tab:ParetoMechErrors})}. Although the formulations are different, both direct strategies and \ac{LoLHR} acquired frontiers that contained similar solutions to the \ac{SO} solution from \citep{Dubourg2011} on average using a comparable number of samples.

\begin{table}
	\centering
	\footnotesize
	\caption{Marginal distributions of the model variables}\label{tab:ex3dists}
	\begin{tabularx}{0.65\linewidth}{Xlll}
		\toprule
		Name & Distribution  & Unit & Description \\
		\midrule
		$M_1$ & $\mathrm{Lognorm.}\left(250 \cdot  10^6,  0.3 \mu_{M_1} \right)$ & $[\mathrm{N}\cdot\mathrm{mm}]$ & Bending moment\\
		$M_2$ & $\mathrm{Lognorm.}\left(125 \cdot  10^6, 0.3 \mu_{M_2} \right)$ & $[\mathrm{N}\cdot\mathrm{mm}]$ & Bending moment\\
		$F_{\mathrm{ax}}$ & $\mathrm{Lognorm.}\left(2.5 \cdot  10^6, 0.2 \mu_{F_{\mathrm{ax}}} \right)$ & $[\mathrm{N}]$ &  Axial Force\\
		$R$ & $\mathrm{Lognorm.}\left(40, 0.1 \mu_{R} \right)$ & $[\mathrm{MPa}]$ & Yield stress \\
		$B$ & $\mathcal{N}\left(\mu_B, 0.01 \mu_B \right)$ & $[\mathrm{mm}]$ & Profile width\\
		$H$ & $\mathcal{N}\left(\mu_H, 0.01 \mu_H\right)$ & $[\mathrm{mm}]$ & Profile height
	\end{tabularx}
	\vspace{-4mm}
\end{table}

\changemine{\ac{SVR} seemed to be a better model choice for this example, as it achieved better results on average both with stationary sampling and \ac{LoLHR} strategies. Moreover, stationary \ac{SVR}  achieved better results compared to \ac{GP} with \ac{LoLHR}. 
In the worst case, \ac{GP} with stationary sampling predicted only unreliable designs, yielding a \ac{HVI} of $0$. This shows the importance of choosing the right model, which may have a higher impact than the used sampling strategy. Therefore, the motivation behind a model agnostic refinement strategy like \ac{LoLHR} has paid off in this example, which seems to be ignored in other proposed strategies so far. In practical applications, the better model can be chosen using the initial data and some quality metric such as the cross-validated error \citepeg{Hastie2009CV} for comparison which does not increase the number of required samples.}

\begin{table}%
	\centering
		\footnotesize
	\caption{\ac{HVI} statistics of each strategy after $10$ runs. Bold numbers represent the best results while \changefinal{the best among the surrogate based solutions are underlined}. Average \ac{SO} performance $\mu_c$ and its standard deviation $\sigma_c$ are computed from the best solution of each Pareto frontier to the \ac{SO} problem in Eq. \ref{eq:ex4orig}. $\tilde{\mu}_h$, $\tilde{\sigma}_h$, $\min_h$, $\max_h$ are same as Table \ref{tab:ParetoAnalErrors}.
	}
	\label{tab:ParetoMechErrors}
	\bgroup
	\def\arraystretch{1.1}
	\begin{tabular}{l|ccccccccc}
		Method              & $\tilde{\mu}_h$      & $\tilde{\sigma}_h$  & $\min_h$             & $\max_h$ & $\tilde{\mu}_\mathcal{F}$ & $\tilde{\mu}_p$ & $\tilde{\mu}_m$ & \changerev{$\mu_{c} / 10^5$} & \changerev{$\sigma_c/ 10^4$}\\
		\hline
		\changerev{Random}  & $302.94$             & $22.49$           & $217.36$               & $345.15$            & $97.1$ & $3.4$ & $1.4\cdot 10^6$ & $2.31$ & $4.86$ \\
		Direct              & $\bm{393.09}$        & $\bm{0.37}$         & $\bm{392.58}$          & $\bm{393.7}$        & $\bm{0}$ & $34.8$ & $9.5 \cdot 10^7$ & $2.15$ & $\bm{0.05}$ \\
		Direct (short)      & $389.46$             & $2.418$             & $385.94$               & $393.27$            & $\bm{0}$ & $\bm{39.5}$ & $9.6 \cdot 10^5$ & $2.16$ & $0.16$\\
		\changerev{Direct (NRBDO)$^1$}  & $277.50$             & -                   & $277.50$               & $277.50$            & $\bm{0}$ & $1$ & $1.9 \cdot 10^7$ & $2.16$ & - \\
		\changemine{\ac{GP} (stat.)}     & $297.26$             & $149.91$            & $0.00$                 & $371.16$            & $258.1$ & $11.9$ & $\underline{\bm{128}}$& $2.70$ & $5.31$\\
		\changemine{\ac{GP} (LoLHR)}     & $307.67$             & $70.21$             & $238.13$               & $392.44$            & $1.8$ & $12.2$ & $\underline{\bm{128}}$ & $2.24$ & $1.21$\\
		\changerev{\ac{GP} (NRBDO)$^1$} & $263.39$             & -                   & $263.39$               & $263.39$            & $\underline{\bm{0}}$ & $1$ & $140$ & $\underline{2.17}$ & - \\
		\changemine{\ac{SVR} (stat.)}    & $329.74$             & $30.43$             & $274.80$               & $391.02$            & $19.6$ & $14.9$ & $\underline{\bm{128}}$ & $2.54$ & $2.15$\\
		\changemine{\ac{SVR} (LoLHR)}    & $\underline{376.45}$ & $\underline{14.27}$ & $\underline{349.83}$ & $\underline{393.04}$& $0.5$ & $\underline{16.2}$ & $\underline{\bm{128}}$ & $2.23$ & $\underline{1.02}$\\
		\changerev{DSA$^1$}             & $288.42$             & -                   & $288.42$               & $288.42$            & $\bm{0}$ & $1$ & - & $2.15$ & -\\
		\changerev{FORM + RIA$^1$}      & $291.64$             & -                   & $291.64$               & $291.64$            & $\bm{0}$ & $1$ & $9472$ & $\bm{2.12}$ & - \\
	\end{tabular}\\
	\egroup
	\begin{flushleft}
		$^1$ as reported by \cite{Dubourg2011}
	\end{flushleft}
\end{table}

Results of all runs are given in Table \ref{tab:ParetoMechErrors}. It can be seen that the direct solutions had a much smaller variance at the cost of very high sample requirements compared to surrogate based strategies. Nonetheless, \ac{LoLHR} results seemed not only to improve the average \ac{HVI}, but also reduced the variance of the results compared to their stationary counterparts. Most \ac{MORRDO} approaches delivered a comparatively good solution to the \ac{SO} problem while also finding solutions for other combinations of objective importance and reliability. \changerev{Moreover, random sampling also seemed to work relatively well on this example. In some cases, it delivered comparable results to the stationary strategies. Models resulting from stationary strategies often yielded a higher number of unreliable subset of solutions due to high approximation error.}

\section{Uncertainty optimization of a lead screw design regarding friction induced vibrations}\label{sec:application}
Lead screws can be used to convert the rotational motion into linear translation. They are often used in elevator and positioning systems. Especially considering acoustic emissions, mechanical solutions are often a better choice compared to hydraulic and pneumatic ones. However, they suffer from some undesired dynamic effects due to frictional forces and stick-slip behaviour, which may cause positioning errors as well as self-induced vibrations. Especially in larger systems, ball screws can be used, which avoid most of these effects. However, ball screws are expensive and it might be infeasible to produce appropriate ball screws depending on the application and material.

An overview of possible use cases and problems can be found in \cite{VahidAraghi2009} and \cite{VahidAraghi2010}. In those works, authors propose control systems for handling the negative effects. However, it may not always be possible or desired to apply control to such mechanisms, for example if they are driven by human force such as in some medicine deployment devices. Even for the applications where control can be applied, finding a design that reduces the probability of such negative behaviour may be advantageous as it reduces the requirements imposed on the controller, thus the total cost and the energy consumption for proper operation. Note that it is possible to find operational points, where these effects do not occur, if the system is analysed deterministically. Nonetheless, introducing uncertainty to design and operational variables according to the real world data may show that even such designs can suffer under the said negative effects. Hence, uncertainty optimization as proposed in this work can be used to find Pareto optimal designs that are robust and reliable.

\subsection{Description of the physical model}

Differential equations of the lead screw model used in this work are given in Chapter 3.5 in \cite{VahidAraghi2009}. In this model, a flexible coupling between the threads and the nut is assumed. The coupling or compliance is modelled as a Kelvin-Voigt element with stiffness and damping coefficients $k_c$ and $c_c$, respectively. In practice, the parameters $k_c$ and $c_c$ of this model depend on material properties, production tolerances as well as finishing methods used on the threads. As a real world data set was not available for inferring the relationship, these parameters are later used directly as optimization variables. 

\changerev{The differential equation of motion for this system can be given in matrix form as}
\begin{equation}
\detm{M}_s\ddot{\detv{y}} + \detm{C}_s\dot{\detv{y}} + \detm{K}_s\detv{y} = \detv{f}_e(\detv{y}, \dot{\detv{y}}) \label{eq:leadscrewode}
\end{equation}
where $\detv{y} = [y_1, y_2]^T$ are relative motions of the lead screw and the nut respectively as derived in \cite{VahidAraghi2009} Chapter 6. The matrices $\detm{M}_s, \detm{C}_s, \detm{K}_s$ are given as follows
\begin{equation}
\begin{gathered}
\detm{M}_s = \begin{pmatrix}
j_m & 0 \\
0 & \hat{m}_l
\end{pmatrix} \\
\detm{K}_s = \begin{pmatrix}
k_\theta + \hat{k}_c(1 - a_2) & \hat{k}_c (-1 + a_2) \\
\hat{k}_c(-1 - a_3) & \hat{k}_c(1 + a_3)
\end{pmatrix} \\
\detm{C}_s = \begin{pmatrix}
c_\theta + \hat{c}_c(1 - a_2) & \hat{c}_c (-1 + a_2) \\
\hat{c}_c(-1 - a_3) & \hat{c}_c(1 + a_3)
\end{pmatrix}
\end{gathered}
\end{equation}
where
\begin{equation}
\begin{gathered}
a_2 =\tau_0 \cot(\lambda) \\
a_3 = \tau_0 \tan(\lambda)
\end{gathered}\label{eq:efftau}
\end{equation}
$j_m$ is the moment of inertia of the screw, $k_\theta$ and $c_\theta$ are the spring and the damping coefficients between the input $\theta_i$ and the screw $\theta$ motion respectively. $\tau_0$ is friction coefficient if no sliding occurs and $\lambda$ is the thread angle. Furthermore, transformed parameters $\hat{m}_l$, $\hat{k}_c$ and $\hat{c}_c$ are given as
\begin{equation}
\begin{gathered}
\hat{m}_l = r_m \tan^2(\lambda) m_l\\
\hat{k}_c = r_m \sin^2(\lambda) k_c \\
\hat{c}_c = r_m \sin^2(\lambda) c_c
\end{gathered}
\end{equation}
where $m_l$ is the effective mass of the nut and $r_m$ is the radius of the screw.

\changerev{Excitation force $\detv{f}_e$ in Eq. \ref{eq:leadscrewode} due to relative motion $\detv{y}, \dot{\detv{y}}$ is modelled in the form
\begin{equation}
\detv{f}_e(\detv{y}, \dot{\detv{y}}) = \left(\tau_0 -  \tau \right) \left( \hat{k}_c (y_2 - y_1) + \hat{c}_c(\dot{y}_2 - \dot{y}_1) + \changefinal{\hat{f}_r} \right) \begin{pmatrix} 
\cot{\lambda} \\
\tan{\lambda}
\end{pmatrix} \label{eq:frictionforce}
\end{equation}
where the transformed force $\changefinal{\hat{f}_r}$ is computed from the axial force $f_r$ as
\begin{equation}
\changefinal{\hat{f}_r} = \frac{r_m \tan(\lambda)}{1 + \tau_0 \tan(\lambda)} f_r
\end{equation}}

$\tau$ in Eq. \ref{eq:frictionforce} is the velocity dependent friction coefficient. A combination of three models (Coulomb, Stribeck and viscous friction) is used to model the frictional behaviour between the threads and the nut. 
Coulomb friction model assumes a constant friction coefficient $\tau_1$, whereas the friction force depends on the velocity $\dot{\theta}$ in Stribeck ($\tau_2$) and viscous ($\tau_3$) models. The magnitude of the friction coefficient $\hat{\tau}$ is given as
\begin{equation}
\begin{gathered}
\hat{\tau} =  \tau_1 + \tau_2 e^{-r_0 |\dot{\theta}|} + \tau_3 |\dot{\theta}| \\
r_0 = \frac{r_m}{v_0\cos(\lambda)}
\end{gathered}\label{eq:frictionmodel}
\end{equation}
where $v_0$ is the so called Stribeck constant that controls the velocity range of the Stribeck effect and \changerev{$\dot{\theta}$ denotes the rotational velocity of the screw in radians:
\begin{equation}
\dot{\theta} = \dot{y}_1 + \dot{\theta}_i
\end{equation}
Thus, the friction coefficient $\tau_0$ in Eq. \ref{eq:efftau} for the case of no relative motion $y_1=0$, \ie $\theta=\theta_i$ can be computed using Eq. \ref{eq:frictionmodel}.}

The direction of frictional force depends on the input velocity $\dot{\theta}_i$ as well as the normal force $f_n$. Thus, the friction coefficient $\tau$ between the threads and the nut can be written as
\begin{equation}
\tau = \hat{\tau} \mathrm{sgn}(\dot{\theta} f_n)
\end{equation}
The normal force $f_n$ between the lead screw threads and the nut is given as
\begin{equation}
\begin{gathered}
f_n = k_c \delta + c_c \dot{\delta} \\
\delta = \changefinal{z} \cos(\lambda) - r_m \theta \sin(\lambda)
\end{gathered}
\end{equation}
\changerev{where \changefinal{$z$} is the nut position and $\delta$ is the relative displacement between the screw and the nut and given as:
\begin{equation}
\changefinal{z} = r_m \tan(\lambda) \left(y_2 + \theta_i \right) + u_{20} \label{eq:pracexbiasterms}
\end{equation}
where $u_{20}$ (Eq. 6.17 in \cite{VahidAraghi2009}) is a constant, used to ensure $y_2=0$ if no relative motion is present. A more detailed description of these equations and their derivations can be found in \cite{VahidAraghi2009}.}

Note that the system is homogenous ($\detv{f}_e(\detv{y}, \dot{\detv{y}}) = [0, 0]^T$), either if the coefficient of friction does not depend on the velocity $\tau = \tau_0$  or if no sliding occurs $y_2=y_1=\dot{y}_2=\dot{y}_1=0$. Experimental measurements in \cite{VahidAraghi2009} show that friction induced vibrations occur in reality meaning $\detv{f}_e(\detv{y}, \dot{\detv{y}}) \neq [0, 0]^T$. Thus, for breaking the symmetry $\dot{y}_2(0)=\dot{y}_1(0)$ in the simulations, the initial value problem for the solution of Eq. \ref{eq:leadscrewode} is defined, as
\begin{equation}
\begin{gathered}
\frac{d}{dt}\begin{pmatrix}\detv{y}(t) \\
\dot{\detv{y}}(t) \\
\end{pmatrix} = \begin{pmatrix} \dot{\detv{y}}(t)\\
\ddot{\detv{y}}(t) \\
\end{pmatrix} \\
\dot{y}_1(0) = \dot{\theta}_0 \\
\ddot{y}_1(0) = \dot{y}_2(0) = \ddot{y}_2(0)=0
\end{gathered}\label{eq:LeadScrewIVP}
\end{equation}
where $\dot{\theta}_0$ is a stochastic variable defining a small initial angular sliding velocity. Since the direction of $f_r$ is positive, $\dot{\theta}_0$ is assumed to be negative.

\subsection{Characteristic vibrations}
Although it looks fairly simple, the non-linearity of $\detv{f}(\detv{y}, \dot{\detv{y}})$ results in \changeown{distinct vibrational behaviours, which cause the negative effects mentioned above.}. Before defining the optimization problem, some of the solutions of Eq. \ref{eq:leadscrewode} to specific parameter combinations are investigated. Input angular velocity of the absolute system was set to $\dot{\theta}_i=100 \; \mathrm{rad} \cdot \mathrm{s}^{-1} \approx 15.9 \; \mathrm{Hz}$ and the initial condition of the relative angular velocity was set to $\dot{\theta}_0=1.375 \cdot{10}^{-3} \mathrm{rad} \cdot \mathrm{s}^{-1}$. The initial condition \changeown{in Eq. \ref{eq:LeadScrewIVP}} was handled as an impulse excitation with the amplitude equal to $\dot{\theta}_0$. Other parameters are given for the investigated combinations in Table \ref{tab:prelimcombs}. Friction parameters are assumed \changeown{to be} constant as given in Table \ref{tab:oscillatorinputs}.

For finding the first eigenfrequency, frequency response spectrum (\ac{FRF}) of the response $\dot{y}_2$ is used. Excitation signal is constructed as $\dot{\theta}_e(t>0) = 0$ except at the initial step where $\dot{\theta}_e(0) = \dot{\theta}_0$. The \ac{FRF} $h_1(\omega)$ is defined as
\begin{equation}
h_1(\omega) = \frac{s_{er}(\omega)}{s_{ee}(\omega)}
\end{equation}
where $s_{er}(\omega)$ is the cross spectral density of the excitation $e(\omega)$ and the response $r(\omega)$ in the frequency domain and $s_{ee}(\omega)$ is the auto spectral density of $e$. Here, each response signal $r(\omega)$ is multiplied by the Hanning window $\tilde{r}(\omega) = w(i) r_i$ and the frequency response is computed as:
\begin{equation}
\tilde{h}_1(\omega) = \frac{2}{\sum_i^{m_t} w(i)} \frac{s_{e\tilde{r}}(\omega)}{s_{ee}(\omega)}
\end{equation}

\begin{figure}[H]
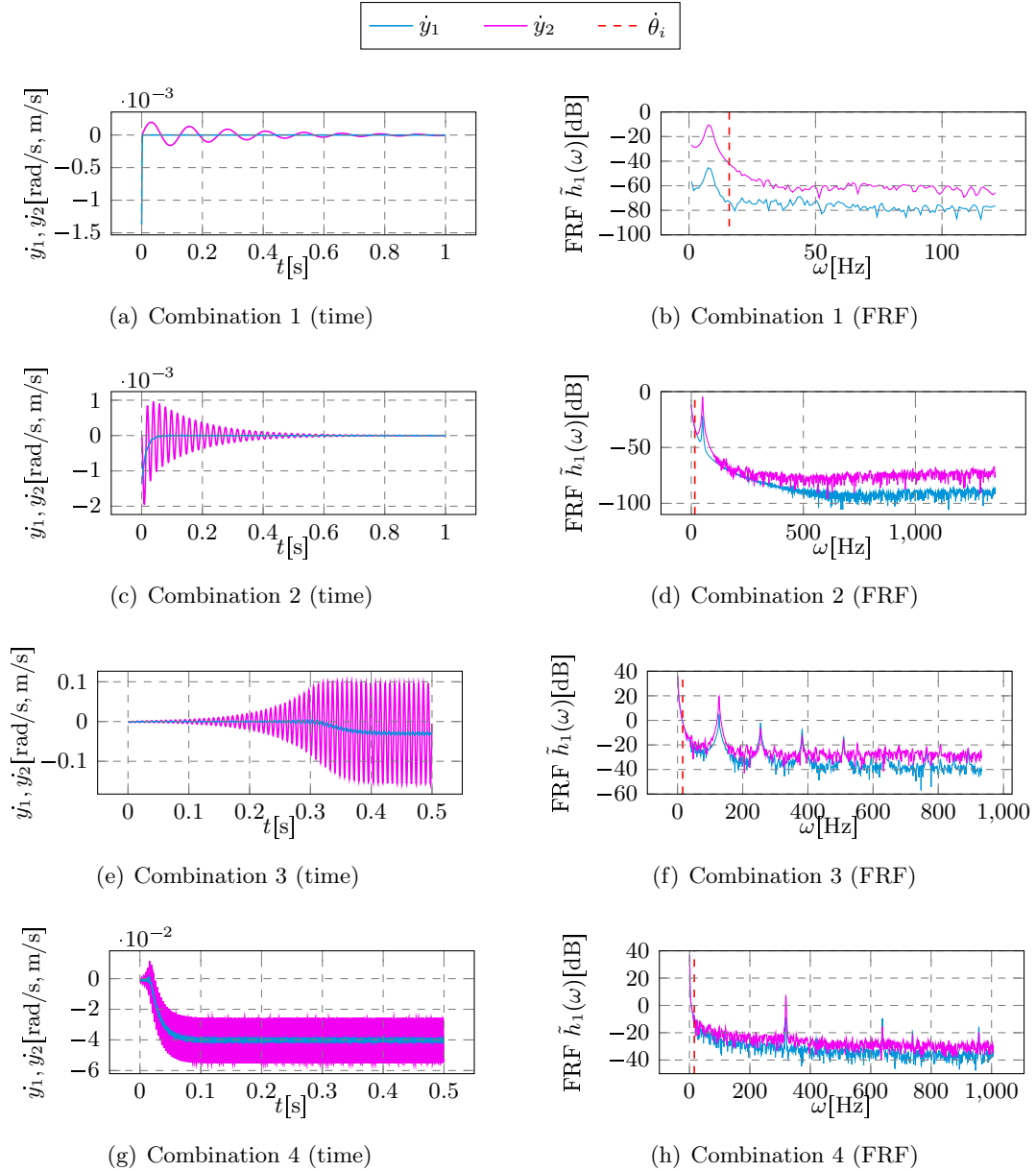

	\centering
	\setlength\fheight{0.15\textheight} 
	\setlength\fwidth{0.4\textwidth}
	\subfigure{%
\ifusepdf%
	\includegraphics[]{Manuscript-figure\thepdffigno.pdf}%
	\refstepcounter{pdffigno}%
\else%
	\input{./Pictures/Practical/prelim/Legend.tex}%
\fi%
} \\
	\addtocounter{subfigure}{-1}
	\subfigure[Combination 1 (time) \label{fig:PracComb1timeresp}]{
\ifusepdf%
	\includegraphics[]{Manuscript-figure\thepdffigno.pdf}%
	\refstepcounter{pdffigno}%
\else%
	\input{./Pictures/Practical/prelim/timedom0.tex}%
\fi%

	}\qquad 
	\subfigure[Combination 1 (FRF) \label{fig:PracComb1freqresp}]{
\ifusepdf%
	\includegraphics[]{Manuscript-figure\thepdffigno.pdf}%
	\refstepcounter{pdffigno}%
\else%
	\input{./Pictures/Practical/prelim/freqdom0.tex}%
\fi%

	}\\
	\subfigure[Combination 2 (time) \label{fig:PracComb2timeresp}]{
\ifusepdf%
	\includegraphics[]{Manuscript-figure\thepdffigno.pdf}%
	\refstepcounter{pdffigno}%
\else%
	\input{./Pictures/Practical/prelim/timedom8.tex}%
\fi%

	}\qquad 
	\subfigure[Combination 2 (FRF) \label{fig:PracComb2freqresp}]{
\ifusepdf%
	\includegraphics[]{Manuscript-figure\thepdffigno.pdf}%
	\refstepcounter{pdffigno}%
\else%
	\input{./Pictures/Practical/prelim/freqdom8.tex}%
\fi%

	}
	\subfigure[Combination 3 (time) \label{fig:PracComb3timeresp}]{
\ifusepdf%
	\includegraphics[]{Manuscript-figure\thepdffigno.pdf}%
	\refstepcounter{pdffigno}%
\else%
	\input{./Pictures/Practical/prelim/timedom12.tex}%
\fi%

	}\qquad 
	\subfigure[Combination 3 (FRF) \label{fig:PracComb3freqresp}]{
\ifusepdf%
	\includegraphics[]{Manuscript-figure\thepdffigno.pdf}%
	\refstepcounter{pdffigno}%
\else%
	\input{./Pictures/Practical/prelim/freqdom12.tex}%
\fi%

	}
	\subfigure[Combination 4 (time) \label{fig:PracComb4timeresp}]{
\ifusepdf%
	\includegraphics[]{Manuscript-figure\thepdffigno.pdf}%
	\refstepcounter{pdffigno}%
\else%
	\input{./Pictures/Practical/prelim/timedom16.tex}%
\fi%

	}\qquad 
	\subfigure[Combination 4 (FRF) \label{fig:PracComb4freqresp}]{
\ifusepdf%
	\includegraphics[]{Manuscript-figure\thepdffigno.pdf}%
	\refstepcounter{pdffigno}%
\else%
	\input{./Pictures/Practical/prelim/freqdom16.tex}%
\fi%

	}
	\caption{Vibrational responses for the parameter combinations}\label{fig:PracComb1resp}
\end{figure}



 Due to Shannon-Nyquist sampling theorem, the maximum limit of the frequency axis was computed as $\omega_{\mathrm{max}} = \frac{1}{2 \Delta t}$ where $\Delta t$ denotes size or the duration of the discrete time steps as given by the differential equation solver. Thus for the Fourier transformation as well as for the following \ac{FRF} analysis, the frequency axis was limited to [0, $\omega_{\mathrm{max}}$] Hz and the rest of the resulting signals with higher frequency were ignored. \textit{solve\_ivp} from \cite{SciPy} was chosen as solver, which adaptively chooses a varying step size to suit the problem, the mean step size was chosen as
\begin{equation}
\Delta t = \frac{2}{m_t - 1} \sum_{i=1}^{m_t - 1} t_{i+1} - t_{i}
\end{equation}
and the excitation as well as the response were resampled after the \changeown{interpolation} to the regular time grid with equidistant step size $\Delta t$. Resampling was necessary since the algorithm used to compute the Fourier transform from \cite{SciPy} only \changeown{accepts} equidistant samples \ie a constant sampling frequency.

At the lower boundary (combination 1), the amplitude of the vibrations are very small (Figures \ref{fig:PracComb1timeresp},\ref{fig:PracComb1freqresp}), which is a desired property for such a design. However, the eigenfrequency of the first mode is relatively low. This is less desired, as $\dot{\theta}_i$ could accelerate and decelerate through this frequency. This may cause negative effects such as mode coupling if sliding occurs at this time, which would result in increased vibration amplitudes and is generally an undesired property as described in \cite{VahidAraghi2009, VahidAraghi2010}.

Increasing the values of the model parameters also seems to increase the first eigenfrequency. Thus, the vibrations depicted in Figures \ref{fig:PracComb2timeresp},\ref{fig:PracComb2freqresp} are more desirable despite the larger amplitude. Nonetheless, the parameter uncertainty may yield worse results and is not accounted \changeown{for} here yet as only the nominal design is depicted. Furthermore, increasing parameter values too much may have other negative consequences such as \textit{negative damping} and \textit{self-locking}.

\begin{table}[H]
	\centering
	\footnotesize
	\caption{Investigated parameter combinations. Units are same as given in Table \ref{tab:oscillatorinputs}} \label{tab:prelimcombs}
	\begin{tabularx}{\linewidth}{XXXXX}
		\toprule
		Notation  & Combination 1 & Combination 2 & Combination 3 & Combination 4\\
		\midrule
		$r_m$ & $1.30 \cdot 10^{-3}$ & $4.57 \cdot 10^{-3}$ & $6.21 \cdot 10^{-3}$ & $7.85 \cdot 10^{-3}$\\
		$\lambda$ & $ 2.0 $ & $ 4.5$ & $ 5.8 $ & $7.05$ \\
		$m_l$ & $ 4.0 $ & $ 4.84 $ & $ 5.26 $ & $ 5.68 $\\
		$\log_{10}(j_m)$ & $ -7.0 $ & $ -6.16 $ & $ -5.74 $ & $ -5.32 $\\
		$k_\theta$ & $ 0.25 $ & $ 0.88 $ & $ 1.20 $ & $ 1.51 $ \\
		$c_\theta$ & $3 \cdot 10^{-4}$ & $1.28 \cdot 10^{-2}$ & $1.91 \cdot 10^{-2}$ & $2.53 \cdot 10^{-2}$\\
		$f_r$ & $ 50.0 $ & $ 134.21$ & $ 176.32$ & $ 218.42$\\
		$\log_{10}(k_c)$ & $ 4.0 $ & $ 5.68 $ & $ 6.53 $ & $ 7.37 $\\
		$c_c$ & $ 25.0 $ & $ 88.16 $ & $ 119.74 $ & $ 151.32 $\\
	\end{tabularx}
\end{table}

Negative damping introduces an instability to the dynamic system and causes self excited vibrations with much larger amplitude than the excitation amplitude. This should generally be avoided as it may cause a design failure due to material limits and other criteria. In Figures \ref{fig:PracComb3timeresp},\ref{fig:PracComb3freqresp}, the amplitude of the vibrations in $\dot{y}_2$ increase up to a factor of $\approx 10$ of the input excitation until $0.3$ seconds. The system stabilizes after $0.3$ seconds as the amplitude of $\dot{y}_2$ becomes constant but a mean shift $\mathrm{E}[\dot{y}_1(t > 0.3)] < 0$ occurs in the relative velocity $\dot{y}_1$ representing the rotational screw motion. Since both $\dot{y}_1$ and $\dot{y}_2$ are centred around $0$, such a mean shift results in a large positioning error caused by an energy loss due to high vibrational amplitudes.

If the thread compliance was not modelled by a spring, self-locking effect would cause the system to stop the motion, thus the relative motion would be $\dot{y}_1(t) = \dot{y}_2(t) = -\dot{\theta}_i$. Besides being unsafe from a design point of view, such a system is numerically unstable and the solver used in this work is not able to converge to a solution. Parameters $k_c$ and $c_c$ improve the numerical stability of the investigated system in addition to modelling the compliance compared to the other models proposed in \cite{VahidAraghi2009}. Nonetheless, self-locking can still be observed as a much larger mean shift in both $\dot{y}_1$ and $\dot{y}_2$ as depicted in Figure \ref{fig:PracComb4timeresp},\ref{fig:PracComb4freqresp}. Although the behaviour is not realistic due to the assumption regarding the compliance, it still allows searching for a design while avoiding this behaviour. Both of these effects are further described in Chapter 3 and 5 in \cite{VahidAraghi2009}. Noisier results of \ac{FRF} in Figures \ref{fig:PracComb3timeresp},\ref{fig:PracComb3freqresp}, \ref{fig:PracComb4timeresp} and \ref{fig:PracComb4freqresp} are due to the resampling procedure as describe above as well as numerical noise.

\subsection{Optimization problem}\label{sec:pracoptprob}
\changeown{The} aim of \changeown{a} such lead screw design should be at least to ensure a robust operability and to keep the production costs low. For simplicity, three objectives were chosen to this end. Firstly, the spring constant $k_c$ between the threads should be minimised as it is assumed that high values would increase the production costs. \changerev{Notice the capitalized variable notations in the following due to the introduced uncertainty to model parameters, \eg the deterministic parameter $k_c$ is now treated as the uncertain variable $K_c \sim F_{K_c}$ and thus noted with a capital letter for a consistent notation with the previous sections. Notation of all uncertain variables and their deterministic counterparts are given in Table \ref{tab:oscillatorinputs}.} Moreover, \changeown{the objectives means and variances} should be minimised due to uncertainty. However, since the distribution $F_{K_c}$ is defined in such a way that the variance depends on the mean value (see Table \ref{tab:oscillatorinputs}), only the mean value is used as an objective. The second and third objectives are to maximize the expected eigenfrequency of the first mode while minimizing its variance.

Lack of oscillations, \ie an overdamped system, is desirable. For such systems, the first mode may become undetectable or may have a very high eigenfrequency ($>5000$ Hz), when analysed with \ac{FRF} \changeown{and} the settings described above. If this case occurs during the optimization due to a parameter combination, \ie if the modal analysis does not detect any modes at all or only those with a higher eigenfrequency than $1024$ Hz, the objectives and constraints were evaluated using $\omega_1 = 1024$ Hz, such that $0 \leq \omega_1 \leq 1024$ Hz holds for any design. This has two advantages. Firstly, it avoids very high frequency values, which might act as outliers decreasing the surrogate model quality. Secondly, it encourages the optimizer to prefer overdamped designs over the ones where oscillations prevail. 

Finally, a maximum positioning error of $0.1$ m is chosen regarding the relative n\changeown{u}t motion $y_2$ as the boundary of operability and failure within the $t_{\mathrm{max}}=1$ second simulation duration. Thus, the optimization problem can be noted as 
\begin{equation}
\begin{gathered}
\argmin_{\boldsymbol{\mu}_{\stov{X}}} \mathrm{E}^{\boldsymbol{\mu}_{\stov{X}}}[K_c], -\mathrm{E}^{\boldsymbol{\mu}_{\stov{X}}}[\Omega_1], \mathrm{Var}^{\boldsymbol{\mu}_{\stov{X}}}(\Omega_1) \\
\textit{\footnotesize s.t.} \; P\left( \mathcal{F} \right) \leq 0.01 \\
\boldsymbol{\mu}_{\stov{X}}^l \leq \boldsymbol{\mu}_{\stov{X}} \leq \boldsymbol{\mu}_{\stov{X}}^u
\end{gathered} \label{eq:PracProb}
\end{equation}
\changefinal{where $\stov{X}$ consists of all variables given in Table \ref{tab:oscillatorinputs}.} Note that $\int_0^{t_{\mathrm{max}}} \dot{y}_2(t) dt$ is used instead of $y_2$\changefinal{,} as $y_2$ has a bias term $u_{20}$ for centring $y_2$ to $0$ as given in Eq. \ref{eq:pracexbiasterms}. The integral is computed as a finite sum using the same discretization steps used by the solver. Thus, the limit state function $g(\cdot)$ is given as:
\begin{equation}
g(\detv{x}) =  \sum_{i=0}^{m_t} \dot{y}_2(t_i) \Delta t_i - 0.1 \approx \int_0^{t_{\mathrm{max}}} \dot{y}_2(t) dt - 0.1 \label{eq:PracLSF}
\end{equation}

The distributions of the model parameters as well as the bounds for the optimization or design parameters are given in Table \ref{tab:oscillatorinputs}. The lead screw moment of inertia $j_m$ and the thread spring constant $k_c$ were formulated over their $\log_{10}$ values since the boundaries span a space of several orders of magnitude.

Values of the only deterministic variables are the parameters of the friction model $\tau_1$, $\tau_2$ and $\tau_3$ as taken from Chapter 4 in \cite{VahidAraghi2009}. Reference value for the rest of the variables are also taken from there and the boundaries are obtained using exploratory calculations similar to the ones depicted earlier. In real world applications, most of the bounds are defined over the availability and capability of bought or produced parts as well as other design requirements. Thus, domain expertise in the application area is necessary for choosing meaningful bounds. Similarly, the distribution of the variables should be obtained using experimental data or the vendor specifications. Nonetheless, since the aim of this study is to compare optimization methods, the problem described above is assumed to be sufficiently representative of realistic applications and the assumptions regarding the statistical distributions are not investigated further.

\begin{table}[H]
	\centering
	\footnotesize
	\caption{Marginal distributions of uncertain variables. $\mathcal{N}(\mu, \sigma)$ denotes a Gaussian variable with mean $\mu$ and standard deviation $\sigma$. $\mathcal{U}[l, u]$ denotes a uniform distribution between the lower bound $l$ and the upper bound $u$. Constant values denote that the parameter was assumed to be deterministic (degenerate distribution).} \label{tab:oscillatorinputs}
	\begin{tabularx}{0.85\linewidth}{Xlll}
		\toprule
		Notation & \multirow{2}{*}{Distribution} & \multirow{2}{*}{Bounds ($[\mu_{X_i}^l, \mu_{X_i}^u]$)} & \multirow{2}{*}{Units} \\
		Det. / Uncertain & & & \\
		\midrule
		$r_m$ / $R_m$ & $\mathcal{N}\left(\mu_{R_m}, 0.05\mu_{R_m} \right)$ & $[1.29625, 9.07375] \cdot 10^{-3}$ & $[\mathrm{m}]$ \\
		$\dot{\theta}_i$ / $\dot{\Theta}_i$ & $\mathcal{N}\left(\mu_{\dot{\Theta}_i}, 0.1\mu_{\dot{\Theta}_i} \right)$ & $[25, 175]$ & $[\mathrm{rad} \cdot \mathrm{s}^{-1}]$\\
		$\dot{\theta}_0$ / $\dot{\Theta}_0$ & $\mathcal{U}\left[-2.5 \cdot 10^{-3}, -2.5 \cdot 10^{-4} \right]$ & - &$[\mathrm{rad} \cdot \mathrm{s}^{-1}]$ \\
		$\lambda$ / $\Lambda$ & $\mathcal{N}\left(\mu_\Lambda, 0.05\mu_\Lambda\right)$ & $[2, 8]$ & $[\mathrm{rad}]$\\
		$m_l$ / $M_l$ & $\mathcal{U}\left[4, 6\right]$ & - & $[\mathrm{kg}]$\\
		$\log_{10}(j_m)$ / $J^*_m$ & $\mathcal{N}\left(\mu_{J^*_m}, 0.1\mu_{J^*_m}\right)$ & $[-7, -5]$ & $\log_{10}\left([\mathrm{kg} \cdot \mathrm{m}^2]\right)$\\
		$k_\theta$ / $K_\theta$ & $\mathcal{N}\left(\mu_{K_\theta}, 0.05\mu_{K_\theta}\right)$ & $[0.25, 1.75]$ & $[\mathrm{N} \cdot \mathrm{m} \cdot \mathrm{rad}^{-1}]$\\
		$c_\theta$ / $C_\theta$ & $\mathcal{N}\left(\mu_{C_\theta}, 0.05\mu_{C_\theta}\right)$ &$[3 \cdot 10^{-4}, 3 \cdot 10^{-2}]$ & $[\mathrm{N} \cdot \mathrm{m} \cdot \mathrm{s} \cdot \mathrm{rad}^{-1}]$\\
		$f_r$ / $F_r$  & $\mathcal{U}\left[50, 250\right]$ & - & $[\mathrm{N}]$\\
		$\log_{10}(k_c)$ / $K^*_c$ & $\mathcal{N}\left(\mu_{K^*_c}, 0.05\mu_{K^*_c}\right)$ & $[4, 8]$ & $\log_{10}\left([\mathrm{N} \cdot \mathrm{m}^{-1}]\right)$\\
		$c_c$ / $C_c$ & $\mathcal{N}\left(\mu_{C_c}, 0.05\mu_{C_c}\right)$ & $[25, 175]$ & $[\mathrm{kg} \cdot \mathrm{s}^{-1}]$\\
		$\tau_1$ / - & $0.218$ & -  & - \\
		$\tau_2$ / - & $2.03 \cdot 10^{-2}$ & - & -\\ 
		$\tau_3$ / - & $-4.47 \cdot 10^{-4}$ & - & $[\mathrm{s} \cdot \mathrm{rad}^{-1}]$ \\
	\end{tabularx}
\end{table}

\subsection{Results}
\begin{figure}[ht]
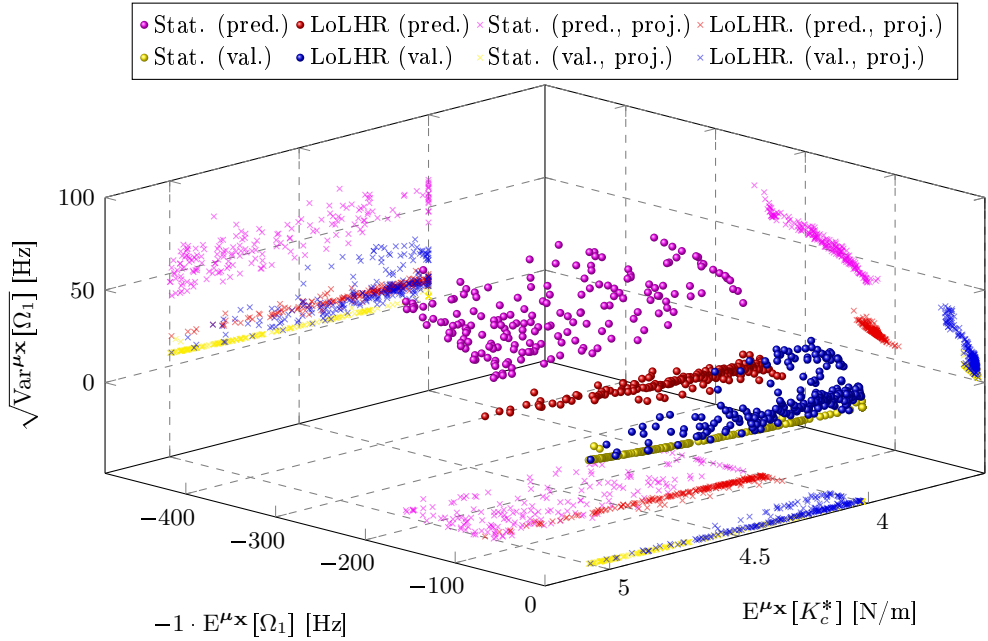

	\centering
	\setlength\fheight{0.5\textwidth} 
	\setlength\fwidth{0.8\textwidth}
\ifusepdf%
	\includegraphics[]{Manuscript-figure\thepdffigno.pdf}%
	\refstepcounter{pdffigno}%
\else%
	\input{./Pictures/Practical/results/Pareto.tex}%
\fi%

	\caption{Predicted and validated Pareto frontiers.}\label{fig:PracPareto}
\end{figure}

Due to the computational burden, only two surrogate based methods were used for the solution. Furthermore, only a portion of the predicted frontiers was selected for validation after the optimization, for which $\mathrm{E}^{\boldsymbol{\mu}_{\stov{X}}}[K_c^*] \leq 5  \; \log_{10}\left([\mathrm{N} \cdot \mathrm{m}^{-1}]\right)$ holds. 
Only these results are presented in the following. \ac{GP} was used both for the stationary sampling and for \ac{LoLHR} \changemine{as it achieved a smaller cross-validation error on the initial data set compared to \ac{SVR}}. The sample budget was set to $256$. For \ac{LoLHR}, the initial data set consisted of $128$ samples and $32$ more were generated at each step. The resulting Pareto frontiers are depicted in Figure \ref{fig:PracPareto}. Standard deviation is displayed instead of the variance for a better visualisation. Furthermore, the negative of the expected eigenfrequency $\mathrm{E}^{\boldsymbol{\mu}_{\stov{X}}}[\Omega_1]$ is displayed to represent each objective as a minimization. Thus, the points that lie further away from the view point in any direction represent designs with better corresponding objective values.

It can be seen that both strategies result in high approximation errors as the predicted and the validated frontiers have large differences, since $256$ samples may not be sufficient to solve this problem with high accuracy regarding the model for $\Omega_1$. However, small data settings are common in industrial applications. The validated results of \ac{LoLHR} deviate less from the predicted values compared to the stationary sampling. Furthermore, the model for the limit state function $g(\cdot)$ seems to be more accurate as all predicted designs by both methods seem to fulfil the probabilistic constraint in Eq. \ref{eq:PracProb}. Detailed results of these frontiers as well as the approximation errors are given in Table \ref{tab:ParetoPracRes}.

\begin{figure}[ht]
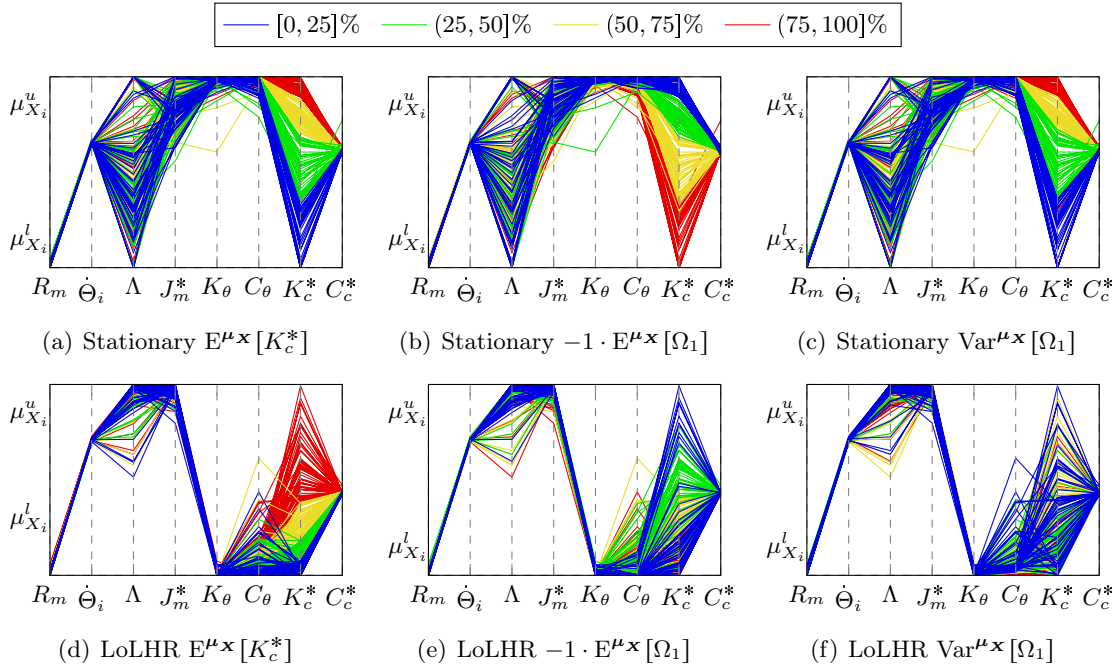

	\centering
	\setlength\fheight{0.25\textwidth}
	\setlength\fwidth{0.33\textwidth}
	\subfigure{%
\ifusepdf%
	\includegraphics[]{Manuscript-figure\thepdffigno.pdf}%
	\refstepcounter{pdffigno}%
\else%
	\input{./Pictures/Practical/results/paracoord/Legend}%
\fi%
} \\
	\addtocounter{subfigure}{-1}
	\subfigure[{Stationary $\mathrm{E}^{\boldsymbol{\mu}_{\stov{X}}}[K^*_c]$} \label{fig:ParaCoordEkcstat}]{
\ifusepdf%
	\includegraphics[]{Manuscript-figure\thepdffigno.pdf}%
	\refstepcounter{pdffigno}%
\else%
	\input{./Pictures/Practical/results/paracoord/ekc_stat}%
\fi%
}
	\subfigure[{Stationary $-1 \cdot \mathrm{E}^{\boldsymbol{\mu}_{\stov{X}}}[\Omega_1]$} \label{fig:ParaCoordEomegastat}]{
\ifusepdf%
	\includegraphics[]{Manuscript-figure\thepdffigno.pdf}%
	\refstepcounter{pdffigno}%
\else%
	\input{./Pictures/Practical/results/paracoord/eomega_stat}%
\fi%
}
	\subfigure[{Stationary $\mathrm{Var}^{\boldsymbol{\mu}_{\stov{X}}}[\Omega_1]$}\label{fig:ParaCoordVaromegastat}]{
\ifusepdf%
	\includegraphics[]{Manuscript-figure\thepdffigno.pdf}%
	\refstepcounter{pdffigno}%
\else%
	\input{./Pictures/Practical/results/paracoord/varomega_stat}%
\fi%
} \\
	\subfigure[{LoLHR $\mathrm{E}^{\boldsymbol{\mu}_{\stov{X}}}[K^*_c]$} \label{fig:ParaCoordEkcadapt}]{
\ifusepdf%
	\includegraphics[]{Manuscript-figure\thepdffigno.pdf}%
	\refstepcounter{pdffigno}%
\else%
	\input{./Pictures/Practical/results/paracoord/ekc_adapt}%
\fi%
}
	\subfigure[{LoLHR $-1 \cdot \mathrm{E}^{\boldsymbol{\mu}_{\stov{X}}}[\Omega_1]$} \label{fig:ParaCoordEomegaadapt}]{
\ifusepdf%
	\includegraphics[]{Manuscript-figure\thepdffigno.pdf}%
	\refstepcounter{pdffigno}%
\else%
	\input{./Pictures/Practical/results/paracoord/eomega_adapt}%
\fi%
}
	\subfigure[{LoLHR $\mathrm{Var}^{\boldsymbol{\mu}_{\stov{X}}}[\Omega_1]$} \label{fig:ParaCoordVaromegaadapt}]{
\ifusepdf%
	\includegraphics[]{Manuscript-figure\thepdffigno.pdf}%
	\refstepcounter{pdffigno}%
\else%
	\input{./Pictures/Practical/results/paracoord/varomega_adapt}%
\fi%
}
	\caption{Pareto designs grouped according to their objective values. \changefinal{$x$-axis represents the mean value of the corresponding parameter.} Blue lines represent the best designs regarding each objective and red ones are the worst. Lower and upper bounds for $\mu_{K^*_c}$ were set to $[4,5]$ for visualisation purposes.}
	\label{fig:ParaCoordDesigns}
\end{figure}

Since the design space is $8$ dimensional, resulting designs are represented as parallel coordinate plots in Figure \ref{fig:ParaCoordDesigns}. It can be seen that the Pareto designs lie in a relatively small portion of the design space for most Variables. Moreover, the proposed designs are quite different for both approaches. Clearly, $\mathrm{E}^{\boldsymbol{\mu}_{\stov{X}}}[K^*_c]$ has a monotonic relationship with the design variable $\mu_{K^*_c}$. Stationary solutions depict an almost monotonic relationship between the mean compliance stiffness $\mu_{K^*_c}$ and other objectives. Increasing $\mathrm{Var}^{\boldsymbol{\mu}_{\stov{X}}}[\Omega_1]$ and $\mathrm{E}^{\boldsymbol{\mu}_{\stov{X}}}[\Omega_1]$ (or decreasing $-1 \cdot \mathrm{E}^{\boldsymbol{\mu}_{\stov{X}}}[\Omega_1]$) seems to require a higher stiffness value $\mu_{K^*_c}$. Furthermore, the values of the thread angle $\mu_\lambda$ span the whole optimization space in this direction. Values for the support stiffness $\mu_{k_\theta}$ and damping $\mu_{c_\theta}$ are set close to upper bound.

In contrast, solutions of \ac{LoLHR} do not represent a monotonic relationship between $\mu_{K^*_c}$ and $\mathrm{E}^{\boldsymbol{\mu}_{\stov{X}}}[\Omega_1]$ or $\mathrm{Var}^{\boldsymbol{\mu}_{\stov{X}}}[\Omega_1]$. This may seem unintuitive but it is partly due to setting the eigenfrequency of overdamped or high frequency vibrations to $1024$ Hz as explained above, for which the eigenfrequency of the first mode $\Omega_1$ was not detected within $[0, \omega_{\mathrm{max}}]$.  Furthermore, values for the support stiffness $\mu_{K_\theta}$ and damping $\mu_{C_\theta}$ are set close to upper bound and the values for the inertia $\mu_{J^*_m}$ as well as the thread angle $\mu_\Lambda$ are set close to the upper bound. This is also visible by the relatively high variance $\mathrm{Var}^{\boldsymbol{\mu}_{\stov{X}}}[\Omega_1]$ in Figure \ref{fig:PracPareto}. Nonetheless, both solutions seem to agree regarding the screw diameter $\mu_{R_m}$, the input velocity $\mu_{\dot{\Theta}_i}$ and the thread damping \changeown{coefficient} $\mu_{C_c}$ which are almost constant for all designs.

A summary of the comparison is given in Table \ref{tab:ParetoPracRes}. Standard deviation  $\sqrt{\mathrm{Var}^{\boldsymbol{\mu}_{\stov{X}}}[\Omega_1])}$ was used for the computation of the \ac{HVI} to reduce the bias of this objective, as the objective $\sqrt{\mathrm{Var}^{\boldsymbol{\mu}_{\stov{X}}}[\Omega_1])}$ is much larger compared to other objectives. This increases the bias towards this objective. The \ac{HVI} results using $\mathrm{Var}^{\boldsymbol{\mu}_{\stov{X}}}[\Omega_1])$ showed a much larger difference in favour of \ac{LoLHR}. Reference was chosen as $(\mathrm{E}^{\boldsymbol{\mu}_{\stov{X}}}[K^*_c], \mathrm{E}^{\boldsymbol{\mu}_{\stov{X}}}[\Omega_1], \sqrt{\mathrm{Var}^{\boldsymbol{\mu}_{\stov{X}}}[\Omega_1])} = [5, 0, 50]$. None of the methods resulted in designs, that turned out to be unreliable during validation ($m_\mathcal{F}=0$). Root mean squared error (\ac{RMSE}) for the first objective $\mathrm{E}^{\boldsymbol{\mu}_{\stov{X}}}[K^*_c]$ is omitted as a surrogate model was not used for its evaluation.

\begin{table}[H]%
	\centering
	\footnotesize
	\caption{Comparison of stationary sampling and \ac{LoLHR}. $p$ denotes the number of Pareto designs and $m_\mathcal{F}$ the number of unreliable designs after validation. $256$ samples i.e. evaluations of the differential equations were used.}
	\label{tab:ParetoPracRes}
	\begin{tabular}{l | c c c c c}
		Method & \ac{HVI} & $p$ & $m_\mathcal{F}$ & \ac{RMSE} $\mathrm{E}^{\boldsymbol{\mu}_{\stov{X}}}[\omega_1]$ & \ac{RMSE} $\mathrm{Var}^{\boldsymbol{\mu}_{\stov{X}}}[\omega_1]$\\
		\hline
		\ac{GP} (stat.) & $1327.99$ & $187$ & $\bm{0}$ & $166.90$ & $2921.09$ \\
		\ac{GP} (\ac{LoLHR}) & $\bm{4015.86}$ & $\bm{225}$ & $\bm{0}$ & $\bm{105.67}$ & $\bm{347.84}$ \\
	\end{tabular}
\end{table}

Using \ac{LoLHR}, Pareto optimal designs with similar $k^*_c$ could be found as given in Combinations $1$ and $2$ from Table \ref{tab:prelimcombs} wrt. the optimization problem given in Eq. \ref{eq:PracProb} considering the design parameter uncertainties given in Table \ref{tab:oscillatorinputs}. In contrast to Combination 2 in Table \ref{tab:prelimcombs}, the Pareto optimal designs were reliable regarding the maximum relative positioning error and in contrast to Combination 1, the oscillations had on average higher first eigenfrequencies. It must be noted that this averaging was biased, as systems with very small oscillations were evaluated to have the first eigenfrequency $\omega_1=1024$ Hz, for which the first mode could not be detected using \ac{FRF} within the frequency interval $[0, 1024]$ Hz.

Finally, to demonstrate the difference between the results regarding the possible gain, one could compare a single design from each frontier. In practice, the designer must choose a design as the final result. If the designer were to choose designs, for which 
\begin{equation}
\mathrm{E}^{\boldsymbol{\mu}_{\stov{X}}}[-\Omega_1] > 2 \pi \dot{\theta}_i\label{eq:PracChoiceCrit}
\end{equation}
holds, a large part of the Pareto frontier predicted by \ac{LoLHR} would be eligible. This is the case for the designs represented by the blue lines in Figure \ref{fig:ParaCoordEomegaadapt}. For some of these designs, $4 \leq \mu_{K^*_c} \leq 4.01 \; \log_{10}\left([\mathrm{N} \cdot \mathrm{m}^{-1}]\right)$ holds.  In contrast, only a few designs from the Pareto frontier acquired by the stationary strategy fulfil this criterion, which have larger values for $\mu_{K^*_c}$. These are represented by the red lines in Figure \ref{fig:ParaCoordEkcstat} and blue lines in Figure \ref{fig:ParaCoordEomegastat} that are close to the displayed upper bound $5 \geq \mu_{K^*_c} \geq 4.9 \; \log_{10}\left([\mathrm{N} \cdot \mathrm{m}^{-1}]\right)$. Thus, almost an order of magnitude could be gained nominally regarding this objective if LoLHR is used instead of the stationary strategy since $\mu_{K^*_c}$ is logarithmic. In any case, sufficiently reliable designs could be found regarding the limit state condition in Eq. \ref{eq:PracLSF} using both strategies.

\section{Conclusion}\label{sec:conc}

\changerev{Local Latin hypercube refinement (\ac{LoLHR}) is proposed for solving \ac{MORRDO} problems to improve design robustness and reliability along with other performance indicators. \ac{LoLHR} increased the sample efficiency compared to strategies without surrogate models and improved the accuracy of the results while decreasing their variance among multiple runs compared to other surrogate based methods, specifically one-shot sampling and other sequential sampling strategies.}

\changerev{Moreover, due to the sparse number of publications considering \ac{MORRDO} problems using surrogate models, \ac{LoLHR} was compared to other surrogate based sequential sampling strategies from the \ac{SORBDO} literature. \ac{LoLHR} achieved comparable results and sample efficiency for the solution of a \ac{SORBDO} benchmark problem while solving the underlying \ac{MO} formulation. For the same computational cost, \ac{LoLHR} was able to find a good approximation of the Pareto frontier, the set of solutions with optimal trade-off values between the reliability and the performance, which contained the solution for the SO problem. This example showcases the motivation for the model-agnostic formulation of \ac{LoLHR} as strategies using \ac{SVR} achieved better performance in this example compared to strategies using \ac{GP}, including the compared work.}

\changerev{More importantly, it has been shown that the sample requirements for \ac{MORRDO} problems could be reduced so far, that optimizing reliability and robustness become almost as affordable as the deterministic solution. The only burden is the validation of the results, for which efficient strategies from the reliability analysis literature can be adopted.}

\changerev{A point left out of the discussion so far is the scalability which is importance for practical applications. For users with access to high-performance computing clusters, it is often more efficient to compute multiple simulations in parallel. \ac{LoLHR} naturally allows proposing multiple points at once. The same cannot be said for probabilistic strategies using \ac{GP} such as the compared method in the second example, which suggests a single point at a time.}

\changerev{Generally, comparative studies are more difficult in \ac{MORRDO} domain, where the resulting Pareto frontiers are either reported visually or only partially recorded at important points. In this work, \ac{HVI} was used as a scalar metric for comparing Pareto frontiers which is recommended for future \ac{MORRDO} research to aid the comparability.}

For the lead screw design, the proposed solutions differed quite a lot when using the stationary sampling and LoLHR. Although reliable designs could be found using stationary sampling, LoLHR achieved better objectives. Especially for the objective regarding the maximization of the eigenfrequency of the first mode, samples generated with LoLHR yielded a model with smaller approximation error especially in the Pareto optimal subspace that contains the Pareto optimal designs, which lead to better solutions.

\changerev{Finally, it can be observed that the current research is mostly interested in developing better strategies for \ac{GP} models. Although \ac{GP} is a powerful model with very useful probabilistic properties, we hope that the results in this work motivate further research on solutions using other surrogate models.}

\section*{Acknowledgement}
The authors would like to express their thanks to the German Federal Ministry of Education
and Research for their support for this project. Support identification no.: 03FH036PX4.%


\bibliography{Literature}

\appendix

\section{Implementation Details}\label{app:impldetails}
The following describes the algorithm specific choices for the sake of completeness and reproducibility. 
\subsection{Supervised learning}\label{app:sml}
Python \citep{PYTHON} library scikit-learn \citep{SKLEARNSHORT} was used for both \ac{GP} and \ac{SVR} as it readily implements all required operations. For \ac{GP}, the sum of squared exponential, rational quadratic and three Matérn kernels ($\nu=0.5$, $\nu=1.5$, $\nu=2.5$) is used as the kernel function
\begin{equation}
\begin{gathered}
k(\detv{x},\detv{x}^0_{i}; \boldsymbol{\Theta}^h, \boldsymbol{\sigma}) = \sum_{s=1}^5 k_j (\detv{x},\detv{x}^0_{i}; \boldsymbol{\theta}^h_s,  \sigma_s^2) \\
\boldsymbol{\Theta}^h_{s, :} = \boldsymbol{\theta}^h_s = \left[\theta^h_{s, 1}, \dots, \theta^h_{s, n}\right] \qquad  \boldsymbol{\sigma}^2 = \left[\sigma^2_1, \dots, \sigma^2_s\right]
\end{gathered}
\end{equation}
which was convenient to implement using scikit-learn. \changerev{A separate length scale vector $\boldsymbol{\theta}^h_j \in \mathbb{R}^{n}$  and variance $\sigma_j^2 \in \mathbb{R}$ were defined for each kernel, which were optimized during the training.} Process noise was set to a small constant ($\sigma_\epsilon=10^{-10}$) to improve the training stability.

For \ac{SVR}, the squared exponential kernel was used, \ie radial basis function in scikit-learn. For obtaining $\lambda$, $\epsilon$ and the kernel parameter $\theta^h \in \mathbb{R}$ for \ac{SVR}, Bayesian optimization \citep{Mockus1994} is used as implemented in the scikit-optimize library \citep{SCIKITOPTSHORT}. Cross-validated mean absolute error was chosen as the objective.
\changemine{Inputs and outputs were normalized to have zero sample mean and unit standard deviation for both strategies.}


\subsection{Unsupervised learning}\label{app:uml}
Scikit-learn also implements \ac{DBSCAN} as used in this work. For the application of \ac{DBSCAN} to LoLHR, labelling points as noise is not desirable as even clusters with low density correspond to some portion of the region of interest. Thus for using \ac{DBSCAN} in an automated fashion, $n_k=n+1$ was set to a small constant and $d_{\min}$ is iterated over the percentile values of all pairwise distances in $\detm{X}^*_k$ increasingly. Whenever more than $90 \%$ of all points in $\detm{X}_{\mathbb{F}}$ were assigned to a cluster (\ie not labelled as noise) and the smallest cluster had at least $10 \%$ of all points in $\detm{X}^*_k$, the search for $d_{\min}$  was terminated and the resulting clusters were used, unless the number of clusters were larger than the number of refinement points $m_s$ at each step. If there are multiple clusters, $1$ point is assigned to their budget iteratively in the order of cluster size regarding the number of member points, until the total number of refinement samples $m_s$ is depleted. If no clusters were found, that satisfy these heuristics, all points are treated as a single cluster.

These heuristics are derived empirically and were necessary as \ac{DBSCAN} is not hierarchical and using it in an automated fashion (\ie without user interaction) or scoring the resulting clusters is not trivial. Nonetheless, using the hierarchical variant of \ac{DBSCAN} (OPTICS \citep{Ankerst1999}) was not preferred as it resulted in too many points to be classified as noise during the preliminary tests. A better solution may exist for the required clustering task depending on the problem, that could improve the results of \ac{LoLHR} especially regarding the variance between different runs. \ac{LoLHR} does not limit the choice of clustering algorithm but note that the number of points to cluster $n_{\mathbb{F}}$ is often high.

\subsection{Uncertainty Quantification}\label{app:ra}
Custom implementations of MC and DS were carried out. When using DS, two hyperparameters need to be set; the number of directions $m_d$ and the maximum search radius $r_{\mathrm{max}}$ in each direction. $m_d$ is given separately for each example. $r_{\mathrm{max}}$ is computed using the target $P^t(\mathcal{F})$ as
\begin{equation}
r_{\mathrm{max}}^2 = F^{-1}_{\chi^2_n}\left(\frac{P^t(\mathcal{F})}{100}\right) \label{eq:Dsrmax}
\end{equation}
where $F^{-1}_{\chi^2_n}$ denotes the inverse CDF of a chi-squared random variable as used for the computation of failure probability in \ac{DS}.  \changerev{Using \ac{DS} with the chosen $r_{\mathrm{max}}$, the smallest possible non-zero probability $\Delta P(\mathcal{F})$ is given as
\begin{equation}
\Delta P(\mathcal{F}) = \frac{P^t(\mathcal{F})}{100 \cdot m_d}
\end{equation}
However, DS can have high bias when predicting $P(\mathcal{F}) = \Delta P(\mathcal{F})$, \ie when only $1$ out of $m_d$ directions contribute to the estimation. Therefore, $r_{\mathrm{max}}$ was chosen as given in Eq. \ref{eq:Dsrmax} for a more accurate estimation of probabilities close to $P^t(\mathcal{F})$.}

Fekete points were used for higher dimensional problems as proposed in \cite{Nie2000}. Brent's method \citep{Brent1971} was used for finding the limit state in each direction as implemented in \cite{SciPy}. For MC, the only hyperparameter is the total number of samples as no early convergence criterion was used. This was chosen according to $P^t(\mathcal{F})$ and is given separately in each example where MC was used.

\subsection{Optimization}\label{app:opti}
\ac{NSGA}-II implementation in \cite{inspyred} was used to solve MO optimization problems. An \ac{LHS} was used as the initial population. \ac{NSGA}-II does not explicitly handle constraints. Often, \changerev{penalty} functions are used to incorporate the constraints to the objectives by worsening the fitness of infeasible designs. In this work, the following \changerev{penalty} function was used
\begin{equation}
\xi_i\left(\detv{x}\right) = \frac{f_i(\detv{x})}{|\overline{f}_i|} -  100 \frac{\min\left(0, \, P^t(\mathcal{F}) - P(\mathcal{F}) \right)}{P^t(\mathcal{F})} \label{eq:nsgaPenRDO}
\end{equation}
where $\overline{f}_i$ is the average of the objective $f_i$ as calculated from the initial population. Note that the scaling with $|\overline{f}_i|$ is also carried out for feasible designs, such that the penalty function in Eq. \ref{eq:nsgaPenRDO} works as intended.


A measure for expressing the quality of fitness of the Pareto frontier in terms of a scalar value was necessary for comparing optimization strategies. Some evolutionary methods \citepeg{Brockhoff2008} use hypervolume indicator (\ac{HVI}) to solve the \ac{MO} problems using \ac{SO} optimizers. Although \ac{NSGA}-II was used instead of those methods in this work, \ac{HVI} is used for the comparison of the resulting frontiers.

\ac{HVI} approximates the hypervolume enclosed by the Pareto frontier with respect to some reference point. 
Note that since the resulting Pareto frontiers are finite sets of points, HVI is computed approximately using the rectangular elements. 
Only Pareto optimal designs that are not worse than the reference point for all objectives, \ie dominating reference, are used for the evaluation. If \ac{HVI} of a frontier is larger, it either has a wider, a denser or a deeper Pareto frontier. The latter assumes the minimization setting. The algorithm proposed by \cite{Fonseca2006} is preferred as an implementation by the authors is publicly available, which is used in this work. 

\end{document}